\documentclass[conference]{IEEEtran}
\usepackage{cite}
\usepackage{amsmath,amssymb,amsfonts}
\usepackage{algorithmic}
\usepackage{graphicx}
\usepackage{textcomp}
\usepackage{xcolor}
\usepackage[hyphens]{url}

\usepackage{comment}
\usepackage{epstopdf}
\usepackage{amssymb}
\usepackage{tikz}
\usepackage{booktabs,caption,fixltx2e}
\usepackage[flushleft]{threeparttable}
\usepackage{dblfloatfix}
\usepackage{multirow}

\usepackage{xcolor}
\definecolor{myBlue}{rgb}{0,0.3,0.5}

\usepackage{xspace}

\def\BibTeX{{\rm B\kern-.05em{\sc i\kern-.025em b}\kern-.08em
    T\kern-.1667em\lower.7ex\hbox{E}\kern-.125emX}}

\title{FlexBlock: A Flexible DNN Training Accelerator with Multi-Mode Block Floating Point Support}
\author{Seock-Hwan Noh*, Jahyun Koo*, Seunghyun Lee*, Jongse Park$^\dagger$, Jaeha Kung*$^\ddag$ \\ *DGIST, $^\dagger$KAIST, Republic of Korea \\ $^\ddag$ Corresponding author}
\begin{document}
\maketitle
\thispagestyle{plain}
\pagestyle{plain}

%%%%%% -- PAPER CONTENT STARTS-- %%%%%%%%

\begin{abstract}

Training deep neural networks (DNNs) is a computationally expensive job, 
which can take weeks or months even with high performance GPUs.
%In conventional training hardware including GPEs and ASICs,
%expensive floating point arithmetic units are used to train the deep learning models.
As a remedy for this challenge, community has started exploring the use of more efficient data representations 
in the training process, e.g., block floating point (BFP).
However, prior work on BFP-based DNN accelerators
rely on a specific BFP representation making them less versatile.
This paper builds upon an algorithmic observation that we can accelerate the training
by leveraging multiple BFP precisions without compromising the finally achieved accuracy.
Backed up by this algorithmic opportunity, 
we develop a flexible DNN training accelerator, 
dubbed FlexBlock, which supports three different BFP precision modes, possibly different among activation, weight, and gradient tensors.
While several prior works proposed such multi-precision support for DNN accelerators, not only do they focus only on the inference, but also their core utilization is suboptimal at a fixed precision and specific layer types when the training is considered.
Instead, FlexBlock is designed in such a way that high core utilization is achievable for i) various layer types, and ii) three BFP precisions by mapping data in a hierarchical manner to its compute units.
%In addition, we tailor the size of blocks sharing an exponent for the target precision level and layer type.
We evaluate the effectiveness of FlexBlock architecture using well-known DNNs on CIFAR, ImageNet and WMT14 datasets.
%To demonstrate the trainability of multi-mode BFP training, we first develop a software training framework 
%building on top of PyTorch, which allows users to configure the block size and precision level of each tensor.
As a result, training in FlexBlock significantly improves the training speed by 1.5$\sim$5.3$\times$ 
and the energy efficiency by 2.4$\sim$7.0$\times$ on average
compared to other training accelerators and incurs marginal accuracy loss compared to full-precision training.

\end{abstract}

\section{Introduction}\label{sec:introduction}

% Talk about deep learning and its importance
With the development of high-performance computing systems and ever-growing open source datasets, 
deep learning has advanced at a very rapid pace.
Due to its accuracy improvement, many applications have started to utilize deep learning including 
computer vision, language modeling, autonomous driving, robotics, and even chip design
~\cite{yolo, gpt3, lin18_autonomous, auto_driving, robotics, chip_design}.
Moreover, many researchers have focused on reducing the model complexity of deep neural networks (DNNs)
to move intelligence into mobile devices~\cite{bnn, han16, 4bit_training, effnet, yoon21, lnpu, mcunet}.
As different deep learning models are actively developed for a wide range of applications and domains, 
various layer types and precision levels are being used.
Unfortunately, there is still a lack of training accelerators optimized for these various conditions with high efficiency.

Generally, deep neural networks are trained in IEEE single-precision format,
i.e., \texttt{FP32}, to minimize the accuracy loss during the training on CPUs/GPUs.
To increase the effective arithmetic and memory bandwidth during the training,
one may reduce the precision in representing activations, weights, and/or gradients~\cite{mixed_precision,cloud-tpu,nnp-t}.
Micikevicius et al. proposed mixed precision training~\cite{mixed_precision},
multiplying two inputs in IEEE half-precision (\texttt{FP16}), 
while accumulating the results in \texttt{FP32}, 
using Tensor Cores in NVIDIA GPUs.
This approach doubles the effective memory bandwidth
and achieves up to 2$\sim$4$\times$ speed-up in DNN training.
%For some applications, such as object detection and speech recognition, 
%loss scaling is used to avoid the divergence during training
%due to the limited dynamic range of \texttt{FP16}.
Instead, one may preserve the exponent bits of \texttt{FP32} (8-bit)
but truncate the mantissa bits to make 16-bit, i.e., \texttt{bfloat16}~\cite{bfloat16}.
There are several commercial DNN training accelerators
that utilize \texttt{bfloat16} to support wider numeric representations~\cite{cloud-tpu,nnp-t}.

Considering the overwhelming size of the recently developed DNNs,
e.g., 469M parameters for AmoebaNet-A~\cite{amoebanet} and 175B parameters in GPT-3~\cite{gpt3},
keeping all tensors in floating point representations would require
huge memory footprint and significant training time.
Recently, a block floating point (BFP) representation has been revived
and applied in training DNNs to improve performance and energy efficiency~\cite{flexpoint, hybrid-bfp}.
However, prior work on BFP-based DNN accelerators rely on a specific BFP representation, making them less versatile and offering limited opportunities for performance and efficiency gains. 
Moreover, DNNs for mobile environment are trained at a low precision
which are suited for the hardware running at that specific precision, 
e.g., \texttt{INT8} on Google Edge TPU~\cite{tpu}.
As training such edge-optimized DNNs entails both low- and high-precision arithmetics, it further motivates the accelerator architecture with the multi-precision support on both fixed- and floating-point representations.
%Thus, an efficient training accelerator needs to have the ability to train DNNs at various precision levels as users may demand a specific precision optimized for their inference hardware.

%With the multi-mode BFP support, this work shows the possibility of training DNNs even at 4-bit mantissa with 8-bit shared exponent by only allowing 8-bit mantissa for computing weight gradients.
%({\it algorithmic contribution}).
%As the demand of DNNs at a reduced precision is increasing (e.g., 1bit, 4-bit or 8-bit),
%training support for variable precision bits becomes necessary.

%% Talk about possible applications of edge servers for distributed computing or federated learning (should we require this context? or simply focus on improving the energy efficiency of the training hardware?)

\begin{figure*}
    \centering
    \includegraphics[scale=0.62]{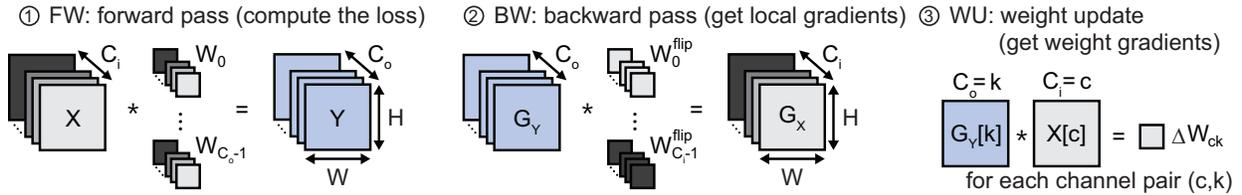}
    \caption{Three important computational steps involved during the training of a convolutional layer.}
    \label{fig:train_process}
\end{figure*}

Unlocking the missing opportunities, we first propose a BFP-based training hardware, dubbed FlexBlock, 
which supports multiple BFP precision modes and layer types. %(\textbf{\textit{hardware contribution}}).
FlexBlock is designed to support 4-bit, 8-bit and 16-bit (sign+mantissas) 
with 8-bit shared exponents\footnote{The basic FlexBlock formats are defined as \texttt{FB12} (=\texttt{FB4+8}) for 4-bit,
\texttt{FB16} for 8-bit, and \texttt{FB24} for 16-bit mantissas with 8-bit shared exponents.}.
With the hardware support, we empirically demonstrate the possibility of training DNNs even with 4-bit arithmetic (\texttt{FB12})
for computing feature maps and local gradients, while allowing 8-bit/16-bit arithmetic (\texttt{FB16/FB24}) for computing weight gradients. %(\textbf{\textit{algorithmic contribution}}).
This aggressive precision scaling during the DNN training results in 5.3$\times$ speedup compared to the training in \texttt{bfloat16} with negligible accuracy loss.
%Compared to recent low-precision training methods using \texttt{FP8}~\cite{fp8,seb_isscc},
%FlexBlock offers more generality (various mantissa bits) and a wider dynamic range (8-bit exponent)
%at a similar or reduced memory bandwidth requirement thanks to block-level exponent sharing\footnote{With the use of \texttt{FB12+WG16} (or basic \texttt{FB12}), we may cut down the required memory bandwidth by 32.7\% (or 49.5\%) over \texttt{FP8}.}.

%memory bandwidth requirement thanks to block-level exponent sharing.

The main contributions can be summarized as follows:
%
%\todo{Jongse}{Should we bring up the fact that we support both INT and FP training or simply argue that
%this is the first work that provides variable-precision training?}
\begin{enumerate}
    \item \textbf{Multi-mode BFP support:} We develop a BFP-based DNN training accelerator supporting the multiple precision modes.
        A DNN model trained in \texttt{FB12}, \texttt{FB16} or \texttt{FB24}
        can be executed on an accelerator supporting \texttt{bfloat16}~\cite{cloud-tpu,nnp-t} 
        or CPUs/GPUs with single-precision as they use the same exponent bits.
        In addition, we can train DNNs in \texttt{INT4}, \texttt{INT8} or \texttt{INT16} with a quantization scaling factor
        since the compute units of FlexBlock are mainly fixed-point arithmetic units.
    \item \textbf{High core utilization:} We maximize the core utilization at all training steps and various layer types by proposing two design techniques:
        i) mapping tensor dimensions in a hierarchical manner to compute units, and
        ii) placing a separate reduction unit for depthwise operations.
        %i) We split the 2-dimensional sub-word parallelism into two 1D sub-word parallelism
        %to reduce the required number of accumulations at a reduced precision.
        %ii) We place a separate reduction unit to improve the core utilization when computing depthwise operations.
    \item \textbf{Low precision training:} We demonstrate the use case of FlexBlock that maximizes the energy efficiency 
        of the training by using 4-bit arithmetics (\texttt{FB12}).
        We accomplish this by statically/dynamically selecting higher bit precisions when computing weight gradients.
\end{enumerate}

\section{Background}\label{sec:background}
\subsection{Training Deep Neural Networks}\label{sec:training}

To train DNN models, three important computational steps are required:
i) computing the training loss ({\it forward pass}; FW), ii) computing local gradients ({\it backward pass}; BW),
and iii) computing weight gradients ({\it weight update}; WU).
As an example, Fig.~\ref{fig:train_process} illustrates these steps for a convolutional (Conv) layer.
We can easily extend the similar analysis to fully-connected (FC) layer as well.
During the forward pass, `$C_i$' input feature maps (fmaps) are convoluted with a set of weight kernels
to generate `$C_o$' output fmaps.
After completing the forward pass, the total loss ($\mathcal{L}$) for a given mini-batch is computed.
Then, the backward pass begins to backpropagate $\mathcal{L}$ to every layer in the network.
In backpropagation, the same convolution operation is performed where the input becomes the local gradient $G_Y=\partial \mathcal{L}/\partial Y$ 
at layer `$l$' and weight kernels are transposed and flipped.
The output of this operation is the local gradient $G_X=\partial \mathcal{L}/\partial X$ for layer `$l-1$'.
At each layer, the weight gradient $\partial \mathcal{L}/\partial W_{ck}$ is computed by performing
a pairwise (and depthwise) full convolution between the local gradient $G_Y[k]$ and input fmap $X[c]$.
The weight gradient is then used to update the weights 
after multiplying it with the learning rate.

\subsection{Block Floating Point}\label{sec:bfp}

As mentioned in Section~\ref{sec:introduction}, DNNs are generally trained with \texttt{FP32}.
The real value $x_i$ in the floating point representation is expressed as
\begin{equation}
    x_i = (-1)^{s_i} \cdot m_i \cdot 2^{e_i},
\end{equation}
where $s_i$ is the sign, $m_i$ is the mantissa, and $e_i$ is the exponent
of the number $x_i$.
Block floating point (BFP) is a special form of the floating point representation,
where a block of $N$ numbers share an exponent corresponding to the number with the largest magnitude~\cite{bfp_system, hybrid-bfp}.
Then, numbers within the block are represented as 
\begin{equation}
    \vec{x} = [x_1, x_2, \ldots, x_N] = [\hat{x_1}, \hat{x_2}, \ldots, \hat{x_N}]\cdot 2^{e_{s}} = \vec{\hat{x}}\cdot 2^{e_{s}}, 
\end{equation}
where $e_s=\lfloor log_2(max(|x_1|,\cdots,|x_N|)) \rfloor$ is the shared exponent
of the block, and $\hat{x_i}=x_i\cdot2^{-e_{s}}$ is the aligned number 
represented by only `sign+mantissa'.
With the BFP representation, therefore, we can perform a dot product in fixed-point arithmetic
without the in-place alignment of intermediate results (cheaper in hardware).
A dot product between two vectors $\vec{w}$ and $\vec{x}$ can be computed by
\begin{equation}\label{eq:bfp_mult}
    \vec{w}\cdot\vec{x} = (\vec{\hat{w}}\cdot\vec{\hat{x}})\cdot 2^{e_w + e_x},
\end{equation}
where $e_w$ and $e_x$ are the shared exponents of $\vec{w}$ and $\vec{x}$, respectively.
One of the goals of this work is to design hardware accelerator that supports various precision levels
in computing $\vec{\hat{w}}\cdot\vec{\hat{x}}$ for the efficient DNN training.

\subsection{Precision-Scalable MAC Array}\label{sec:variable_mac}

% Needs to shorten the text if required
Many research efforts have been made over the recent years 
to design precision-scalable MAC arrays that enable
on-device DNN inference~\cite{wacv16, kung16, envision, dnpu, unpu, bitfusion, bitblade, loom, survey19}.
Earlier work use a simple data gating scheme to zero out operand(s)
to minimize the dynamic power consumed by the MAC array~\cite{wacv16, kung16, dnpu, unpu}.
A bit-serial data fetching on weight tensors has been 
presented to allow fully-variable weight precision (\textit{temporal scalability})~\cite{unpu}.
This temporal scalability has been extended to both operands, i.e., input and
weight tensors, to simplify the computing logic~\cite{loom}.
However, the bit-serial approach consumes varying clock cycles depending on the
precision level and requires more complex control logic.
On the other hand, Shin et al. proposed to utilize sub-word parallelism on weight tensors~\cite{dnpu}.
A full-precision multiplier is built out of multiple sub-multipliers, which are always active (\textit{spatial scalability}).
To provide the sub-word parallelism on both operands (2D parallelism),
a systolic array in which each processing engine consists of sixteen multipliers
has been presented~\cite{bitfusion}.
As pointed out by Camus et al.~\cite{survey19}, the 2D sub-word parallelism shows the best energy efficiency
when designing the precision-scalable MAC array.

\section{Motivation}\label{sec:motivation}

In this paper, we aim to devise a multi-precision support accelerator architecture for DNN training. 
While the existing multi-precision architectures are exclusively designed for inference, one may think that the na\"ive adaptation of such architecture is sufficient for training.  
Thus, we first delve into the prior work and identify the limitations of existing architectures that we target to optimize in this work.
%

\begin{figure}
    \centering
    \includegraphics[scale=0.65]{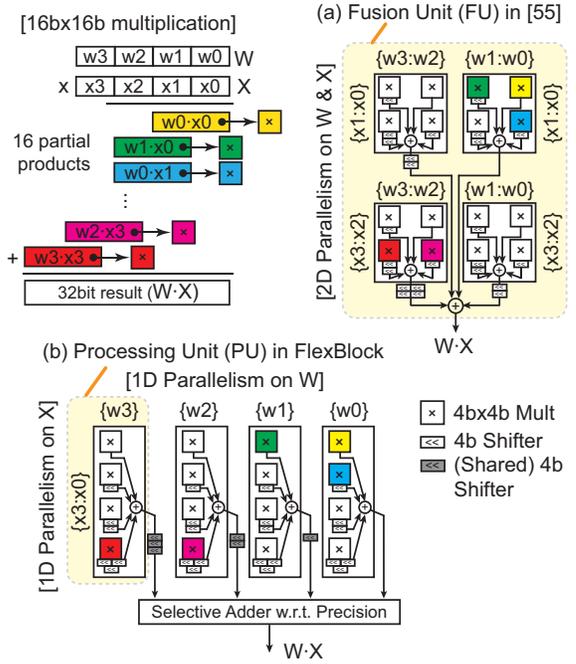}
    \caption{An illustration of performing a 16b$\times$16b multiplication using two precision-scalable multipliers:
    (a) a fusion unit (FU) in Bit Fusion~\cite{bitfusion} and (b) a processing unit (PU) in the proposed FlexBlock.
    Four PUs in FlexBlock split the 2D sub-word parallelism into two 1D sub-word parallelisms.}
    \label{fig:motiv_parallel}
\end{figure}

\begin{figure*}
    \centering
    \includegraphics[scale=0.6]{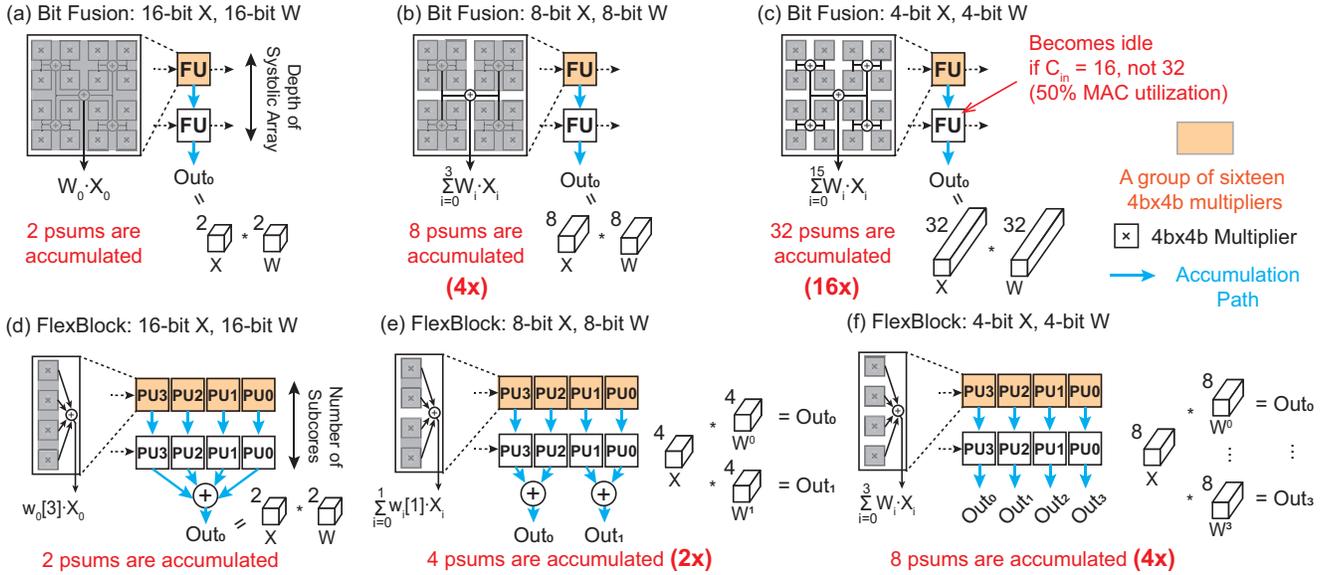}
    \caption{A motivational example explaining the rate of increase in the number of accumulations at a reduced precision: (a-c) Bit Fusion and (d-f) FlexBlock.
    All accumulations within an FU in Bit Fusion need to be completed prior to passing the partial sum.
    However, only a subset of accumulations are completed at each PU in FlexBlock and the remaining accumulations, if required, happen at the end.
    In addition, FlexBlock provides a better input data reuse requiring less number of new operands.}
    \label{fig:mac_compare}\vspace{-2mm}
\end{figure*}

%In this paper, our primary goal is to maximize the MAC utilization when training DNNs 
%at any precision levels and layer types.
{\bf Limitation: }One of the representative works on a precision-scalable MAC array is Bit Fusion~\cite{bitfusion}. 
Fig.~\ref{fig:motiv_parallel}(a) shows a fusion unit (FU) presented in~\cite{bitfusion} using the 2D sub-word parallelism.
Each partial product in a 16b$\times$16b multiplication is computed at a dedicated 4b$\times$4b 
multiplier (some are color-coded).
Since all accumulations within an FU need to be completed prior to passing the result to the next FU,
the number of accumulated partial sums (psums) quadruples 
when there is 2$\times$ precision reduction on both operands (X and W).
A simple motivational example on this issue is provided in Fig.~\ref{fig:mac_compare}(a-c).
This may result in the underutilization of MACs
limiting the speed-up expected by the precision scaling.
In addition, Bit Fusion requires a significant number of shifters, 
e.g., $\sim$98k 4-bit shifters for the 64$\times$64 array.
To improve the power-efficiency, 
BitBlade~\cite{bitblade} clusters multipliers with the identical shift length and reduces
the number of shifters by 93.8\% compared to Bit Fusion.
Still, the number of accumulations increases at the same rate as Bit Fusion with precision scaling.

{\bf Solution: }To mitigate this problem,
a subset of multipliers are grouped together, as a processing unit (PU), to realize 1D sub-word parallelism on X (Fig.~\ref{fig:motiv_parallel}(b)).
Across multiple PUs, i.e., four in FlexBlock, 1D sub-word parallelism on W is realized
where psums from PUs are accumulated at the end depending on the precision mode.
The advantage of splitting the 2D parallelism is more clear by looking at the example shown in Fig.~\ref{fig:mac_compare}(d-f).
Instead of forcing all PUs to perform the same vector multiplication with a lengthy vector dimension,
each accumulation path can be assigned to compute different output channels.
With this hierarchical sub-word parallelism, the number of accumulated psums doubles even with the 2$\times$ precision reduction
on both X and W (Fig.~\ref{fig:mac_compare}(d-f)).

\begin{figure}[t]
    \centering
    \includegraphics[scale=0.74]{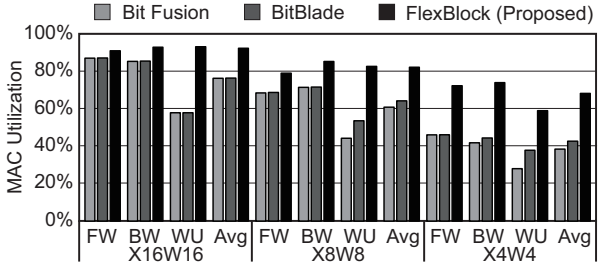}
    \caption{Comparison of the MAC utilization during the training of 
        MobileNetV1 with mini-batch size of 16 using various precision-scalable 
        MAC arrays (Bit Fusion~\cite{bitfusion}, BitBlade~\cite{bitblade}
        and the proposed FlexBlock).}
    \label{fig:compare_util}\vspace{-2mm}
\end{figure}

{\bf Analysis: }To quantitatively examine the implication of such architectural difference, we analyze the MAC utilization of FW, BW and WU steps on Bit Fusion, BitBlade, and our FlexBlock architectures, as we change the input and weight tensor precisions.  
For the analysis, we assume the training variants of Bit Fusion and BitBlade architectures attached with necessary \texttt{FP32} accumulators and BFP modules at the end of the MAC array. 
Fig.~\ref{fig:compare_util} reports that the MAC utilization is 76.5\% for both Bit Fusion and BitBlade
on training MobileNetV1~\cite{mobilenet}
%\footnote{Note that Bit Fusion and BitBlade do not support the training. 
%We assumed each MAC array is supported by block floating point modules.} 
when both input (X) and weight (W) tensors are 16-bit (denoted as X16W16).
%Without the BFP modules which are equipped in FlexBlock, Bit Fusion and BitBlade cannot support the training.
When we reduce the precision to 8-bit (X8W8), the utilization reduces by 13.8\% on average.
If we further reduce the precision from 8-bit to 4-bit (X4W4), additional 22.0\% utilization drop is observed on average.
This is because a much larger number of accumulations are required at a reduced precision to avoid wasting computing resources.

% \usepackage{graphicx}
\begin{table}[t]
\centering
\caption{Examples of two and three-dimensional operations supported by FlexBlock}
\label{tab:op_modes}
\scalebox{1.0}{
\begin{tabular}{|c|c|l|}
\hline
\textbf{ Modes } & \multicolumn{2}{c|}{\textbf{Operations}}                                                                                    \\ \hline\hline
\textbf{2D}     & \multicolumn{2}{c|}{\begin{tabular}[c]{@{}c@{}}Computing $\partial \mathcal{L} / \partial W$, Depthwise Conv, \\Dilated Conv, Up Conv\end{tabular}}          \\ \hline
\textbf{3D}     & \multicolumn{2}{c|}{\begin{tabular}[c]{@{}c@{}}General Conv, Pointwise Conv, FC\\(for both forward and backward pass)\end{tabular}} \\ \hline
\end{tabular}%
}\vspace{-1mm}
\end{table}

%\todo{Jaeha}{Please check if the following makes sense}
This MAC underutilization problem gets exacerbated when considering the weight gradient calculation, i.e., WU step,
since the WU consists of a number of depthwise operations that require a small number of accumulations and thus utilize a small subset of MAC units. 
%The small number of accumulations associated with depthwise operations
%including the weight update exacerbate the underutilization problem
%(refer to `WU' in Fig.~\ref{fig:compare_util}).
As the depthwise operations do not entail computations across multiple channels, we classify them as 2D operations in this paper.
Table~\ref{tab:op_modes} summarizes 2D and 3D DNN operations supported by the FlexBlock core.
%More importantly, the prior work on precision-scalable MAC arrays do not support the training.

\section{Design of a FlexBlock Core}\label{sec:reconfig}

\subsection{Hierarchical Design in FlexBlock}\label{sec:design_hier}

\begin{figure}
    \centering
    \includegraphics[scale=0.68]{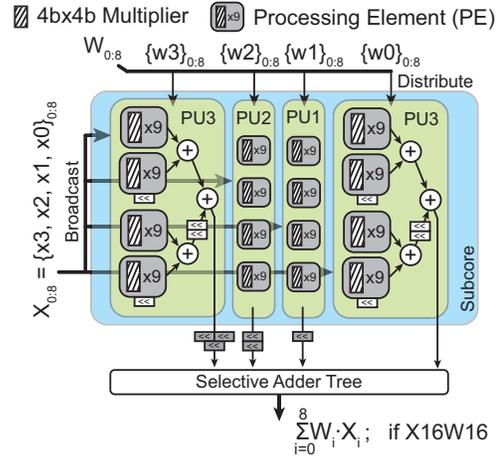}
    \caption{Hierarchical structure of hardware modules in Flex-Block. Four PUs form a subcore in FlexBlock.}
    %For the data distribution, input data are broadcast to PUs and weight parameters are distributed across PUs.}
    \label{fig:flexblock_motiv}\vspace{-2mm}
\end{figure}

For the fine-grained control of hardware modules depending on the precision mode and layer type, 
we designed a processing core of FlexBlock to have a hierarchical structure, i.e.,
multiplier$\rightarrow$processing element (PE)$\rightarrow$processing unit (PU)$\rightarrow$subcore (Fig.~\ref{fig:flexblock_motiv}).
A multiplier in FlexBlock accepts both signed and unsigned operands similar to~\cite{bitfusion}.
One global sign bit is used to indicate whether the input/weight tensor is in signed or unsigned representation.
%Thus, 5-bit signed multipliers are used in FlexBlock (Fig.~\ref{fig:multiplier}).
Then, nine multipliers are clustered together to make one PE to efficiently map and compute 2D convolutions\footnote{Having nine
multipliers removes the burden of \texttt{im2col} operations on the host CPU.
However, the number of multipliers per PE can vary depending on the design strategy (e.g., 8 multipliers per PE).}.
%; Fig~\ref{fig:motiv_parallel}(b) simply shows a PE with one multiplier.
In FlexBlock, the sub-word parallelism on the input tensor X is achieved across four PEs in a PU.
For the 16-bit input tensor X, each 4-bit sub-word (x3, x2, x1 or x0) is mapped to the corresponding PE.
Four PEs are then clustered to form a PU.
Note that a PU in Fig~\ref{fig:motiv_parallel}(b) has a PE with one multiplier for the simple illustration.
Another sub-word parallelism on the weight tensor W is realized across four PUs in a subcore.
For the 16-bit weight tensor, only a 4-bit sub-word (w3) is mapped to PU3 and transferred 
to all PEs within the PU assuming that the precision of the input tensor is 16-bit.
The remaining three sub-words, i.e., w2, w1 and w0, are distributed to the rest of PUs, respectively.

%FlexBlock is efficiently reconfigured
%to maximize the core utilization at various precision levels and layer types.

\subsection{Bit Reconfigurability}\label{sec:bit_reconf}

In FlexBlock, we provide three levels of bit precision, i.e., 4-bit, 8-bit and 16-bit,
for `sign+mantissa' of each tensor for Conv/GEMM operations.
For all three precision levels, we use the same 8-bit shared exponents.
Thus, we define three basic FlexBlock formats as \texttt{FB12} (X4W4), \texttt{FB16} (X8W8) and \texttt{FB24} (X16W16).
We can also mix-and-match different mantissa bits on the associated tensors, 
e.g., 8-bit inputs, 4-bit weights, and 16-bit gradients, allowing $3^3=27$ combinations.
%When designing the core, we focus on maximizing the MAC utilization 
%of PEs at all precision combinations.

\begin{figure}[t]
    \centering
    \includegraphics[scale=0.58]{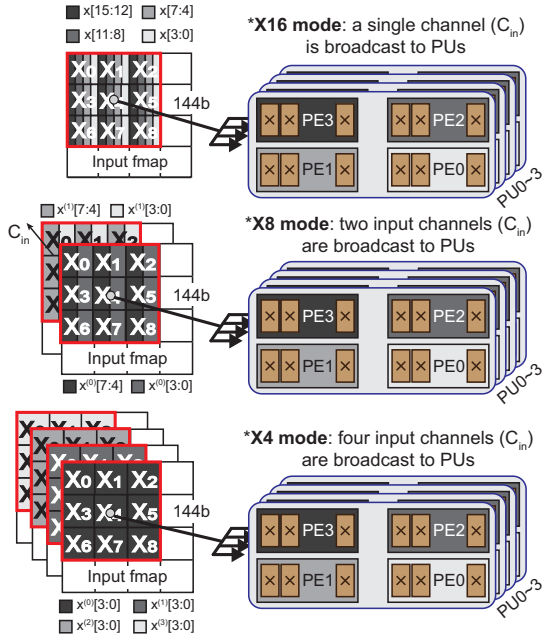}
    \caption{Examples on distributing input data to PEs in each PU
        depending on the input tensor precision (feature maps or local gradients).
        Inputs are broadcast to PUs in a single cycle.}
    \label{fig:i_distribute}
\end{figure}

\textbf{Sub-word parallelism on X:} Fig.~\ref{fig:i_distribute} shows how the input activations (forward pass) or local gradients (backward pass)
are delivered to PEs and PUs for the Conv3 layer\footnote{ConvK represents a K$\times$K convolutional layer.}.
For the 16-bit mode (X16), each fmap element consists of four 4-bit sub-words.
Thus, each sub-word is mapped to a 4b$\times$4b multiplier in the corresponding PE with the matching gray color.
Note that input operands are {\it broadcast} to PUs rather than mapping different
input channels to each PU.
For the 8-bit mode (X8), each fmap element consists of two 4-bit sub-words.
In this case, we can broadcast two input channels to PUs with the 144-bit interconnect.
Similarly, we broadcast four input channels to PUs in the 4-bit mode (X4).
In this case, each input channel is mapped to a PE in the PU.

\begin{figure}[t]
    \centering
    \includegraphics[scale=0.58]{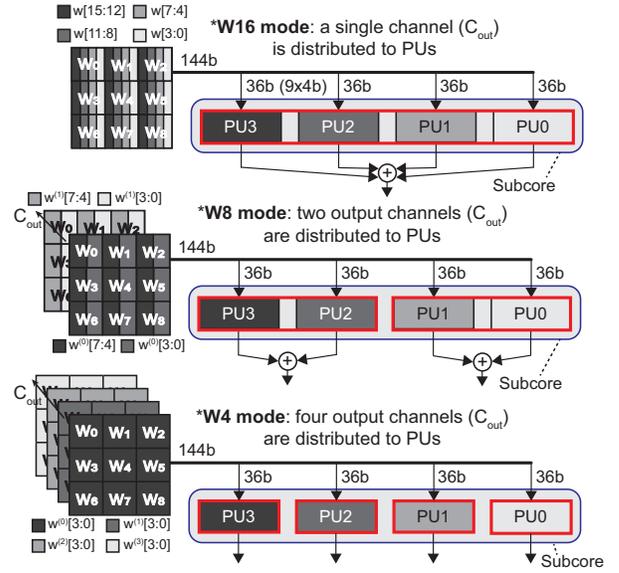}
    \caption{Examples on distributing weight data to processing units in a subcore
        depending on the precision of the weight. 
        In this example, 16-bit inputs are assumed 
        (thus, a single input fmap is shared across multiple weight kernels).}
    \label{fig:w_distribute}
\end{figure}

\textbf{Sub-word parallelism on W:} Fig.~\ref{fig:w_distribute} shows how the weight parameters are delivered to PUs for the Conv3 layer as well.
For the 16-bit mode (W16), a 4-bit sub-word of each weight parameter is delivered to the corresponding PU.
The other three sub-words are {\it distributed} to the remaining PUs in a subcore.
%Here, we assume 16-bit input tensors thus a single input fmap is shared by PEs.
In this mode, the outputs from all PUs are accumulated by the selective adder tree.
For the 8-bit mode (W8), we partition PUs into two clusters and assign the dimension $C_{out}$
across clusters.
Thus, PU2-3 and PU0-1 provide partial sums for $C_{out}=k$ and $C_{out}=k+1$, respectively.
The selective 4-way adder tree at the end of the reduction unit
produces the two partial sums.
For the 4-bit mode (W4), four output channels are distributed to four PUs.
In this case, each PU produces a partial sum for the assigned output channel 
bypassing the selective adder tree.
%The utilization of output bandwidth varies by the precision combinations 
%(Fig.~\ref{fig:w_distribute} shows examples with 16-bit input fmaps).

\subsection{Mapping Various DNN Layers}\label{sec:layer_reconf}

\begin{table}[t]
\centering
\caption{Tensor dimensions mapped to each FlexBlock module depending on layer types}
\label{tab:mapped_dim}
\scalebox{0.94}{
\begin{tabular}{|c|c|}
\hline
\textbf{FlexBlock Module}  & \textbf{Mapped Tensor Dimension}                                                                          \\ \hline\hline
\textbf{Subcore}           & \begin{tabular}[c]{@{}c@{}}C\textsubscript{in} (Conv1/FC, Conv3), C\textsubscript{out} (2D Mode),\\ W/H (Conv5, Conv7)\end{tabular} \\ \hline
\textbf{Processing Unit}       & C\textsubscript{out} (determined by the precision of weights)                                                             \\ \hline
\textbf{Processing Element} & C\textsubscript{in} (determined by the precision of inputs)                                                               \\ \hline
\textbf{Multiplier}     & \begin{tabular}[c]{@{}c@{}}C\textsubscript{in} (Conv1/FC),\\ W/H (Conv3, Conv5, Conv7, 2D Mode)\end{tabular}                        \\ \hline
\end{tabular}}
\end{table}

\begin{figure*}[t]
    \centering
    \includegraphics[scale=0.76]{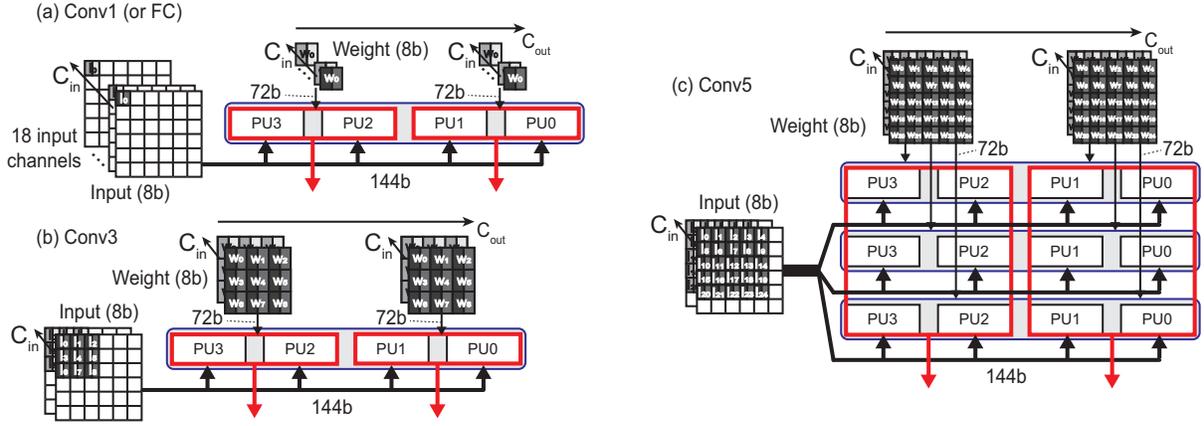}
    \caption{Examples on how FlexBlock groups subcores depending on the layer type and distributes input/weight tensors.
        FlexBlock can efficiently process (a) a Conv1 or fully-connected layer, (b) a Conv3 layer and (c) a Conv5 layer.
        Note that three subcores are grouped together to compute the Conv5 layer. 
        Since we have six subcores in FlexBlock, we can map another set of input feature maps 
        (e.g., two subsequent input channels for 8-bit activations) to the remaining subcores.
        To compute a Conv7 layer, which is omitted here for brevity, FlexBlock groups all six subcores.}
    \label{fig:layer_reconfig}
\end{figure*}

\textbf{Mapping 3D operations:} In this subsection, we present how the input/weight tensors are being mapped to
a FlexBlock core for various DNN layers.
The compute modules in FlexBlock are designed with hierarchy
so that different tensor dimensions can be easily mapped to
these modules depending on the layer type as summarized in Table~\ref{tab:mapped_dim}.
%Note that the only component that reconfigures is the selective adder tree.
%Other than that hardware modules are fixed and only the order in which data are stored sequentially changes.
Some examples on how FlexBlock clusters subcores or PUs depending on the layer type
are shown in Fig.~\ref{fig:layer_reconfig}.
We assume \texttt{FB16} (X8W8) as a precision level for the illustration here.
For the Conv1 or FC layer, the only partial sums to be
accumulated are in dimension $C_{in}$.
Thus, input elements and the corresponding weight parameters across the dimension $C_{in}$ 
are mapped to subcores, PEs and multipliers (Fig.~\ref{fig:layer_reconfig}(a)).
Subcore0 is responsible for the first 18 input channels ($C_{in}$ = 0$\sim$17), Subcore1 is in charge of computing the next 18 input channels ($C_{in}$ = 18$\sim$35), and so on.
For the Conv3 layer, the only difference over the Conv1 case is in mapping operands to multipliers (Fig.~\ref{fig:layer_reconfig}(b)).
The dimension mapped to the multipliers becomes the fmap width/height ($W$/$H$).
With larger weight kernels, e.g., a Conv5 or Conv7 layer,
we cluster multiple subcores to assign all input elements in the $W$/$H$ dimension with the size of a weight kernel.
To maximize the core utilization on various layer types, we placed six subcores in FlexBlock.
For the Conv5 layer, we make two clusters with three subcores each. 
Then, the core utilization becomes $(5\times5)/(3\times 9)=0.93$.
For the Conv7 layer, we make a cluster with all six subcores providing the core utilization of $(7\times7)/(6\times 9)=0.91$.

\begin{figure}[t]
    \centering
    \includegraphics[scale=0.75]{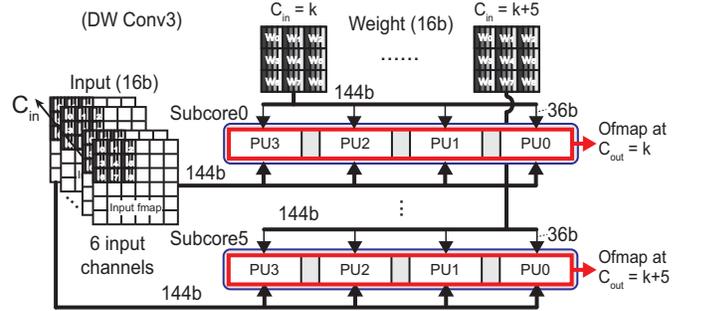}
    \caption{An example of mapping 2D operations to FlexBlock subcores.
        Here, we select a widely used depthwise Conv3 as an example.
        Note that outputs from each subcore move horizontally to the reduction unit for 2D mode.}
    \label{fig:dw_conv}
\end{figure}

% Describe how we map 2D operations on FlexBlock with a simple figure
\textbf{Mapping 2D operations:} Thus far, we explained the mapping strategy for 3D operations
where the outputs from PUs are vertically accumulated 
by the reduction unit for 3D mode.
FlexBlock has a separate reduction unit for 2D operations to maximize the core utilization.
The depthwise convolution (DW Conv) layer is a good example of the 2D operation.
In Fig.~\ref{fig:dw_conv}, we illustrate the mapping of a DW Conv3 layer to FlexBlock subcores.
For 2D operations, we recommend to keep 8-bit or 16-bit for each tensor since some 2D operations are sensitive 
to the precision reduction (see Section~\ref{sec:accuracy}).
Then, all outputs from PUs per subcore are accumulated by the 4-way adder tree 
in the 2D reduction unit.
For the DW Conv3, each output that comes from the subcore is for each output channel $C_{out}$.
We cluster subcores for larger weight kernels, e.g., DW Conv5 (three subcores) or DW Conv7 (six subcores),
which is similar to scaling up the Conv size in the 3D mode (Fig.~\ref{fig:layer_reconfig}(c)).
%Instead of taking the output vertically, however, we route the output horizontally in the 2D mode.

\section{Overall Architecture}\label{sec:overall_arch}

\begin{figure*} 
    \centering
    \includegraphics[scale=0.5]{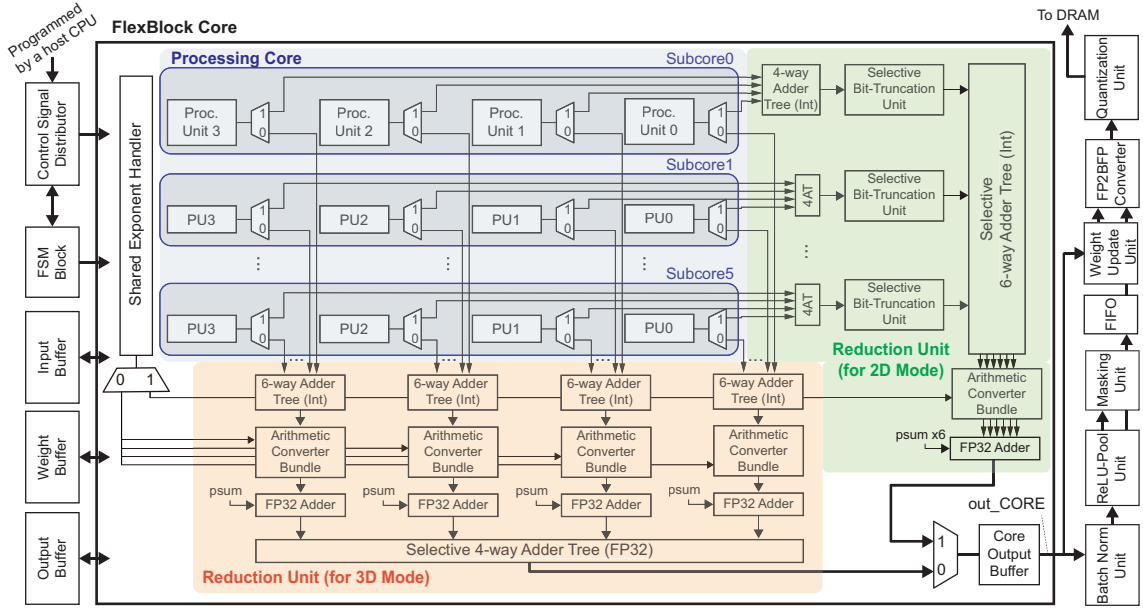}
    \caption{The overall microarchitecture of FlexBlock core. 
        There are two reduction units where each unit is dedicated to either 2D or 3D operation mode.
        Our FlexBlock core is capable of processing various block floating point numbers
        with 8-bit shared exponents (e.g., \texttt{FB12}, \texttt{FB16} and \texttt{FB24}).
        %In addition, FlexBlock allows us to train DNN models in \texttt{INT4}, \texttt{INT8} and \texttt{INT16}
        %with a global scaling factor.
        The `out\_CORE' goes to a weight update or a batch normalization unit.
        Before storing the data to DRAM, we extract shared exponents of sub-tensors at a FP2BFP converter 
        for computing the next layer.}
    \label{fig:overall_arch}
\end{figure*}

% overall architecture of FlexBlock
The detailed microarchitecture of FlexBlock is provided in Fig.~\ref{fig:overall_arch}.
There are three major blocks in the FlexBlock core design: i) a processing core, ii) a reduction unit for 2D operations,
and iii) a reduction unit for 3D operations.
%The processing core handles convolutions or GEMM operations between sub-tensors, i.e., only mantissas with the shared exponent.
%Then, the reduction units accumulate the partial sums while considering the precision level and the layer type.
%Note that FlexBlock is designed to support various DNN layers with high core utilization 
%({\it layer reconfigurability}).
%In addition, we delicately group multipliers so that
%compute modules are not wasted when training DNN models with mixed or reduced precision
%({\it bit reconfigurability}).

\subsection{Major Building Blocks}\label{sec:flexblock_design}

% Explain the details on the design of a FlexBlock core
\subsubsection{Processing Core} As mentioned earlier, 
each processing core consists of six subcores to maximize the core utilization on various DNN layers.
The `sign+mantissas' of input and weight tensors are mapped to these subcores
and the multiplication results are accumulated together by integer adders,
i.e., $(\vec{\hat{w}}\cdot\vec{\hat{x}})$ in Eq.~(\ref{eq:bfp_mult}),
if their shared exponents are extracted a priori.
%Subcores are grouped differently when dealing with different types
%of DNN layers as explained in Section~\ref{sec:layer_reconf}
%to maximize the core utilization.
%Each PU has four PEs and there are nine multipliers at each PU.
%Depending on the tensor type and the bit precision,
%operands are distributed to PEs differently.
%For the 16-bit input tensor, each 4-bit sub-word is mapped to each PU.
%As a notation, we split the 16-bit operand into four 4-bit sub-words 
%and call them \{HH, HL, LH and LL\}.
%Since each PU requires nine operands,
%144-bit (=9$\times$16-bit) per PU is required to fully utilize the entire multipliers.
%For the 16-bit weight tensor, only a 4-bit sub-word (HH) is mapped to PU0 and transferred 
%to all PEs within the PU assuming that the precision of the input tensor is 16-bit.
%Thus, each PU requires 36bit (=9$\times$4-bit) interconnect for the weight distribution.
%The remaining three sub-words, i.e., HL, LH and LL, are distributed to the rest of PEs.
%More details on the distribution of input or weight tensors depending on the precision level
%will be discussed in Section~\ref{sec:bit_reconf}.
Depending on the bit precision (\texttt{FB12}, \texttt{FB16} or \texttt{FB24}),
we block each sub-tensor differently that shares the same exponent.
Table~\ref{tab:block_size} summarizes the minimum block size for each FlexBlock format on various DNN layers.
With the reduced precision, the number of input channels mapped to the processing core
doubles (\texttt{FB16}) or quadruples (\texttt{FB12}) compared to \texttt{FB24} (refer to Fig.~\ref{fig:i_distribute}).
In addition, the number of input channels grouped by the block
proportionally decreases as the size of weight kernels increases.
This fine-grained control of the block size
is proposed to make use of all the multipliers available in the processing core,
i.e., high core utilization, for various precision levels and DNN layers.

% Explain the details on the design of a dual-path reduction unit
\subsubsection{Reduction Units} The prior work~\cite{loom, survey19, bitfusion, bitblade} 
on the design of precision-scalable MAC arrays have
two major limitations: i) no support for DNN training, and ii) low core utilization for 2D operations.
The former is handled by supporting the block floating point arithmetic in FlexBlock with a shared exponent handler,
arithmetic converters placed prior to FP32 accumulation units, and an FP2BFP converter (Fig.~\ref{fig:overall_arch}).
The latter is resolved by placing the dedicated reduction unit for 2D operations
along with the default reduction unit for 3D operations ({\it dual-path reduction units}).
The core utilization for the 2D operation becomes important for the DNN training 
since the computation of a weight gradient ($\Delta \mathbf{W}^l_{ck}$) involves a depthwise full convolution between 
every pair of the local gradient ($\mathbf{G_Y}^{l+1}_k$) and input fmap ($\mathbf{X}^{l}_c$).
Due to the nature of channel-wise computations, 
a small number of multiplication results are required to be accumulated, 
which significantly reduces the core utilization in the prior work (Section~\ref{sec:motivation}).

% Need a table or a figure to explain how we group sub-tensors for BFP computations
\begin{table}[t]
\centering
\caption{The minimum block size of an input tensor sharing the exponent at each FlexBlock format on various layer types}
\label{tab:block_size}
\scalebox{0.9}{
\begin{tabular}{|c|c|c|c|c|c|}
\hline
\textbf{Layer Type}             & \textbf{Format} & \textbf{Block Size} & \textbf{Layer Type}             & \textbf{Format} & \textbf{Block Size} \\ \hline\hline
\multirow{3}{*}{\textbf{CONV1/FC}} & \texttt{FB12}                 & 1$\times$1$\times$216         & \multirow{3}{*}{\textbf{CONV5}} & \texttt{FB12}                 & 5$\times$5$\times$8           \\ \cline{2-3} \cline{5-6} 
                                & \texttt{FB16}                 & 1$\times$1$\times$108         &                                 & \texttt{FB16}                 & 5$\times$5$\times$4           \\ \cline{2-3} \cline{5-6} 
                                & \texttt{FB24}                 & 1$\times$1$\times$54          &                                 & \texttt{FB24}                 & 5$\times$5$\times$2           \\ \hline
\multirow{3}{*}{\textbf{COVN3}} & \texttt{FB12}                 & 3$\times$3$\times$24          & \multirow{3}{*}{\textbf{CONV7}} & \texttt{FB12}                 & 7$\times$7$\times$4           \\ \cline{2-3} \cline{5-6} 
                                & \texttt{FB16}                 & 3$\times$3$\times$12          &                                 & \texttt{FB16}                 & 7$\times$7$\times$2           \\ \cline{2-3} \cline{5-6} 
                                & \texttt{FB24}                 & 3$\times$3$\times$6           &                                 & \texttt{FB24}                 & 7$\times$7$\times$1           \\ \hline
\end{tabular}}
\end{table}

\subsection{Other Functional Blocks for BFP-based DNN Training}
\subsubsection{Batch Normalization Unit}
When training DNNs, batch normalization (BN) is an essential step
to find better weight parameters with faster convergence.
The BN reduces the internal covariate shift making the training process more stable~\cite{tradBN}.
To update the BN parameters, i.e., running mean $\mu$ and variance $\sigma^2$,
all input tensors need to be read from DRAM three times~\cite{bn2019}.
In~\cite{bn2019}, authors present a fusion technique to reduce the read accesses
to twice.
In FlexBlock, we use range batch normalization~\cite{rangeBN} 
that further reduces the number of DRAM accesses to one with simpler hardware modules.
%also requires two DRAM accesses in processing the BN layer with simpler hardware modules.
%Refer to Section~\ref{sec:bn_hardware} for more details.

\subsubsection{ReLU-Pool Unit}

\begin{figure} 
    \centering
    \includegraphics[scale=0.58]{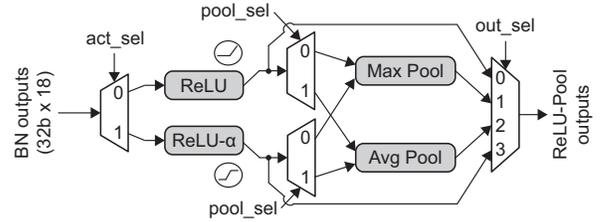}
    \caption{Design of a reconfigurable ReLU-Pool unit placed after batch normalization unit.}
    \label{fig:relu-pool}
\end{figure}

\renewcommand{\thefigure}{13}
\begin{figure*}[b]
    \centering
    \includegraphics[scale=0.72]{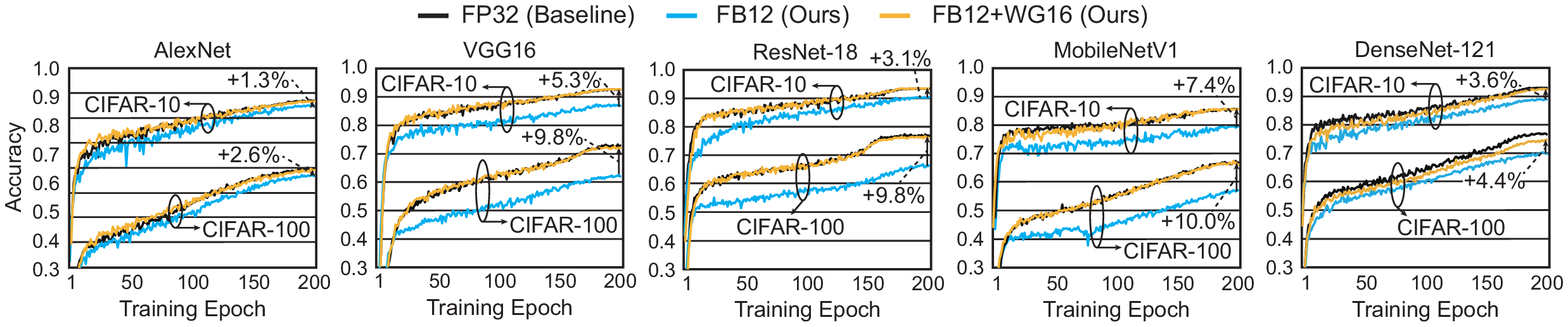}
    \caption{Accuracy of five CNN models trained in \texttt{FB12} on CIFAR datasets improves by setting
        8-bit `sign+mantissas' for weight gradients (WG), i.e., denoted as \texttt{FB12+WG16}.
        With the precise weight gradient computation, the accuracy of FlexBlock closely matches with the baseline (\texttt{FP32})
        even with \texttt{FB12} for most computations.}
    \label{fig:accuracy_imprv}\vspace{-2mm}
\end{figure*}

In general, the BN layer is followed by a nonlinear activation function
and an optional pooling layer in CNNs.
%The layer orders appeared in various CNN models are summarized in Table~\ref{tab:layer_order}.
Since the pooling layer may not exist between the BN layer and Conv layer,
we design a reconfigurable ReLU-Pool unit as shown in Fig.~\ref{fig:relu-pool}.
%\footnote{Also, other deep learning models such as MLPs and Transformers do not have pooling layers.}.
For the activation function, FlexBlock provides ReLU and ReLU-$\alpha$.
The ReLU-$\alpha$ unit accepts a clipping value $\alpha$ as a parameter, which is set to 6 for MobileNets~\cite{mobilenet, mobilenetv2}.
The $\alpha$ can also be trained for improving the accuracy of quantized neural networks~\cite{pact}.
For the pooling layer, FlexBlock allows no pooling, max pooling, or avg pooling
by controlling the `out\_sel' signal.

\renewcommand{\thefigure}{12}
\begin{figure} 
    \centering
    \includegraphics[scale=0.58]{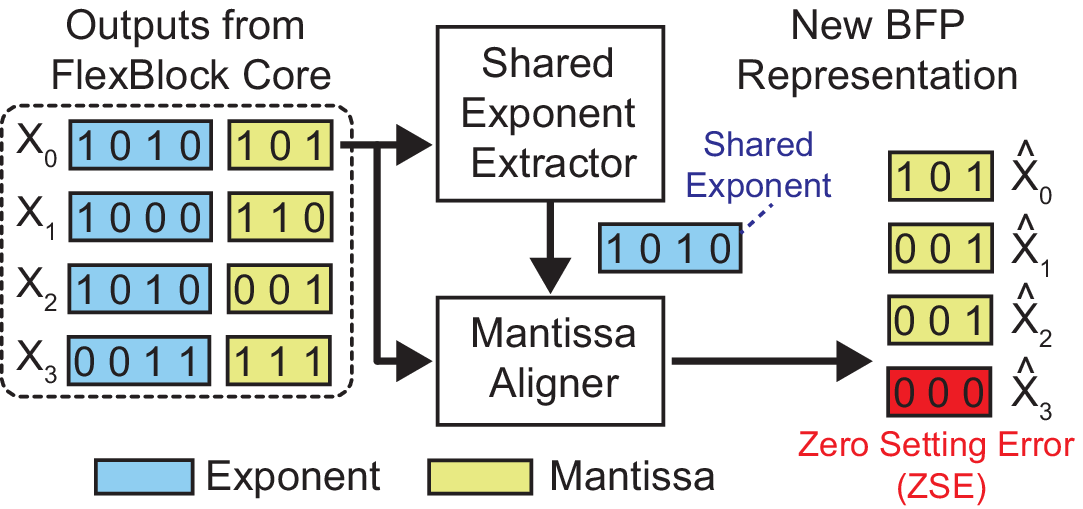}
    \caption{A simple illustration of the process performed by the FP2BFP converter. 
    The number of zero setting errors (ZSEs) can be used to determine the precision mode of each tensor at runtime (Section~\ref{sec:dynamic}).}
    \label{fig:fp2bfp}\vspace{-3mm}
\end{figure}

\begin{comment}
\subsubsection{Masking Unit}

A masking unit is placed right after the ReLU-Pool unit
that extracts a bitmap, which is used during the backward pass
to minimize the energy consumption of unnecessary fmap accesses.
For instance, the backward pass on the ReLU layer
propagates the gradient value only at a location
at which the output feature is non-zero (positive).
As presented in~\cite{gist},
it is possible to store a binarized map of the ReLU output 
if the pooling layer is followed by the ReLU layer.
In FlexBlock, we fuse bitmaps of the ReLU and pooling layers
for more efficient data storage.
%More details are discussed in Section~\ref{sec:reduce_data}.
\end{comment}

\subsubsection{Weight Update Unit}

A weight update unit is directly connected to the core output buffer.
After the weight gradients are computed by the processing core,
they are passed to the weight update unit at which
a vector unit is placed to multiply a learning rate $\eta$ to the weight gradients.
Then, the weight parameters ($W^l_{ck}$) are subtracted by the scaled weight gradients ($\eta\cdot\Delta W^l_{ck}$)
using element-wise subtract units.

\subsubsection{Block Floating Point Converter}

The FP2BFP converter is placed after the weight update unit to prepare
input/gradient and weight tensors for computations at the next layer.
This unit blocks each sub-tensor by the pre-defined block size
as summarized in Table~\ref{tab:block_size} depending on the precision level
and layer type of the following layer.
It has a shared exponent extractor and mantissa aligners that
let the processing core of FlexBlock handle only fixed-point computations (Fig.~\ref{fig:fp2bfp}).
The quantization unit is followed by the FP2BFP unit to minimize communication overheads
between the core and DRAM.

\section{Methodology}\label{sec:methodology}

\subsection{Software Framework for BFP-based DNN Training}

For evaluating the accuracy of a fine-grained blocking of sub-tensors during the DNN training, 
we implemented a configurable BFP trainer using PyTorch.
%\footnote{Our configurable block floating point trainer (BFPsim) 
%will be available at https://github.com/blinded\_for\_review/.}. 
The BFPsim first defines the network, then it replaces all `torch.nn.Conv2d' and `torch.nn.Linear' modules 
with `BFP.Conv2d' and `BFP.Linear' modules in a configuration file provided by the user (`\texttt{bfp\_config.json}'). 
The bit precision of each tensor can be configured in this file as well as the block size that shares the exponent.
To monitor the impact of the reduced precision during the computation of weight gradients,
we allow users to individually control the mantissa bits for local and weight gradients.
%(i.e., `\texttt{wg\_bit(bp)}').
The values for blocked sub-tensors are converted to BFP format by extracting the shared exponent
and aligning mantissas within the block.
Then, we perform pseudo BFP computations in the BFPsim, which means that we store the converted BFP values in \texttt{FP32} 
to fully utilize internal functions of PyTorch.
%After running BFPsim, we can extract some statistics to check how 
%many zero-setting errors have occurred by using the current BFP configuration.
%Also, the value distribution of each tensor can be extracted to check 
%which tensor distribution is far from the baseline due to the precision scaling.

\subsection{Hardware Implementation}
%%% No need to edit this section (Checked)

To evaluate the proposed FlexBlock hardware in detail,
we implemented RTL of all building blocks shown in Fig.~\ref{fig:overall_arch} except SRAMs.
Then, they are synthesized in 65nm CMOS technology using Synopsys Design Compiler (ver. N-2017.09-SP5~\cite{synopsys_dc}) 
running at 333MHz ($T_{clk}=$ 3ns).
To extract more accurate area estimation, post-PnR area is obtained by using Synopsys IC Compiler~\cite{synopsys_icc}.
For the power analysis, we extracted saif files after setting different BFP modes and layer types 
then feeding testbenches to FlexBlock.
The extracted saif files are then used in Design Compiler to estimate the power consumption of FlexBlock
with more realistic switching probabilities at each BFP mode.
%% Comment (Jaeha): Shall we keep this statement and project the area of FlexBlock in 28nm technology??
%We also taped out the first prototype in 28nm CMOS technology which contains only
%FlexBlock core, i.e., processing core + dual-path reduction units.
The energy consumption and cycle time of accessing SRAMs in 65nm are estimated by using CACTI~\cite{cacti, cactip}.
We assume a FlexBlock accelerator with a 512KB input buffer, a 512KB weight buffer, and a 256KB output buffer
that are distributed to 64 FlexBlock cores.
Double buffering is utilized to hide the DRAM access latency when possible.
The 64 FlexBlock cores are capable of computing $54\times64=3,456$ 16b$\times$16b MAC operations in \texttt{FB24}.
The number of operations increases by 4$\times$ (13,824) or 16$\times$ (55,296) 
when the mode is set to \texttt{FB16} or \texttt{FB12}.

\section{Experimental Results}\label{sec:experiments}
\subsection{Accuracy of DNN Training with Multi-Mode BFP Support}\label{sec:accuracy}

\subsubsection{Benchmarks} To evaluate the algorithmic stability of training DNNs in various BFP formats, we selected four datasets, i.e., CIFAR-10, CIFAR-100~\cite{cifar-10}, ImageNet~\cite{imagenet} and WMT14~\cite{wmt14}.
Note that FlexBlock is able to individually set mantissa bits for activations, weights, and gradients to achieve the minimum training cost in terms of energy consumption.
First, we extensively studied the training of five representative CNNs, i.e., AlexNet~\cite{alexnet}, VGG16~\cite{vggnet}, ResNet-18~\cite{resnet},
MobileNetV1~\cite{mobilenet} and DenseNet-121~\cite{densenet} on simple CIFAR datasets (Fig.\ref{fig:accuracy_imprv}).
%This runtime overhead is due to the expensive tensor conversion (\texttt{FP32}-\texttt{BFP}-\texttt{FP32}) between every two DNN layers.
Then, we trained ResNet-18 on ImageNet and Transformer~\cite{transformer} on WMT14 in various BFP formats to check how well they scale to more complex tasks (Fig.~\ref{fig:accuracy_new}).

\subsubsection{Basic BFP Formats} 
As a baseline, we trained all benchmarks in \texttt{FP32}.
As another baseline, we trained the benchmarks using mixed precision supported by Tensor Cores in NVIDIA RTX3090.
In the mixed precision training, multiplications are performed in \texttt{FP16}
and accumulations are done in \texttt{FP32}, i.e., `\texttt{FP16+FP32}' in Table~\ref{tab:accuracy_base}.
First, we trained all benchmarks with basic BFP formats, i.e., \texttt{FB24}, \texttt{FB16} and \texttt{FB12}.
%while other precision combinations can be tested as well.
In the basic BFP format, `sign+mantissa' bits of all tensors are set
to the same bit-width, e.g., 8-bit for activation, weight, and gradient tensors in \texttt{FB16}.
Throughout the experiments, we stick to the block size provided in Table~\ref{tab:block_size}
to evaluate the training/test accuracy of FlexBlock.
If we look at the accuracy comparisons in Table~\ref{tab:accuracy_base}, 
the test accuracy with \texttt{FB24} or \texttt{FB16} is similar to the baselines.
However, the test accuracy is significantly lower than the baselines
when we train the model with \texttt{FB12} (-6.21\% on average for CIFAR datasets and -11.46\% for ImageNet, respectively).
For Transformer trained on WMT14 dataset, \texttt{FB12} still provides similar perplexity to other high-precision
data formats (Fig.~\ref{fig:accuracy_new}).

%% Add figure for ImageNet and WMT14
\setcounter{figure}{13}
\renewcommand{\thefigure}{\arabic{figure}}
\begin{figure}[t]
    \centering
    \includegraphics[scale=0.7]{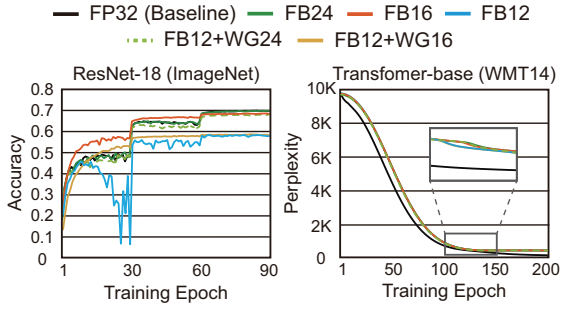}
    \caption{Comparison of training curves between different 
        precision formats on more complex datasets, i.e., ImageNet and WMT14 En-De (best viewed in color).}\vspace{-2mm}
        %Each curve shows the test accuracy during the training process.}
    \label{fig:accuracy_new}
\end{figure}

\begin{table}
\centering
\caption{Achieved final accuracy when using a diverse set of precisions for training DNNs on four well-known datasets}\label{tab:accuracy_base}
\scalebox{0.9}{
\begin{tabular}{|c|c|c|c|c|c|c|}
\hline
\textbf{Precision}     & \textbf{\texttt{FP32}} & \textbf{{\begin{tabular}[c]{@{}c@{}}\texttt{FP16}\\ +\texttt{FP32}\end{tabular}}} & \textbf{\texttt{FB24}} & \textbf{\texttt{FB16}} & \textbf{\texttt{FB12}} & \textbf{{\begin{tabular}[c]{@{}c@{}}\texttt{FB12}\\ +\texttt{WG16}\end{tabular}}} \\ \hline\hline
\textbf{Dataset} & \multicolumn{6}{c|}{\textbf{CIFAR-10 (Top-1 Accuracy)}}\\ \hline
\textbf{AlexNet}        & 87.12 & 86.96 & 87.06 & 86.68 & 85.11 & 86.41 \\ \hline
\textbf{VGG16}          & 92.83 & 92.90 & 92.93 & 92.97 & 87.43 & 92.71 \\ \hline
\textbf{ResNet-18}       & 93.37 & 93.68 & 93.79 & 93.75 & 90.36 & 93.50 \\ \hline
\textbf{MobileNetV1}    & 87.07 & 86.85 & 85.98 & 86.67 & 79.80 & 87.22 \\ \hline
\textbf{DenseNet-121}       & 93.37 & 93.61 & 93.06 & 93.32 & 89.42 & 93.01 \\ \hline\hline
\textbf{Dataset} & \multicolumn{6}{c|}{\textbf{CIFAR-100 (Top-1 Accuracy)}}\\ \hline
\textbf{AlexNet}        & 59.52 & 59.21 & 59.39 & 59.77 & 57.03 & 59.66 \\ \hline
\textbf{VGG16}          & 73.35 & 73.16 & 73.06 & 73.20 & 62.51 & 72.35 \\ \hline
\textbf{ResNet-18}       & 77.26 & 76.88 & 77.45 & 77.23 & 66.69 & 76.45 \\ \hline
\textbf{MobileNetV1}    & 67.40 & 66.80 & 66.95 & 67.60 & 57.11 & 67.06 \\ \hline
\textbf{DenseNet-121}       & 77.24 & 77.57 & 77.68 & 77.05 & 70.53 & 74.93 \\ \hline\hline
\textbf{Dataset} & \multicolumn{6}{c|}{\textbf{ImageNet (Top-1 Accuracy)}}\\ \hline
\textbf{ResNet-18}        & 69.95 & 69.23 & 69.92 & 68.60 & 58.49 & 68.20\dag \\ \hline\hline
\textbf{Dataset} & \multicolumn{6}{c|}{\textbf{WMT14 En-De (Perplexity)}}\\ \hline
\textbf{Transformer-base}        & 3.87 & 3.92 & 4.33 & 4.29 & 4.26 & 4.27 \\ \hline
\end{tabular}%
}\vspace{0.5mm}
\begin{tabular}{l}
\dag For ImageNet dataset, training with \texttt{FB12+WG24} achieves the  similar \\accuracy to the baseline.\\
\end{tabular}
\end{table}
%% 승현학생(TODO): If time permits, let's run FP16+FP32 (not mandatory though)

\subsubsection{BFP Variants} 
The accuracy degradation in \texttt{FB12} is due to the limited precision by having 4-bit `sign+mantissas'.
Note that all FlexBlock formats use 8-bit shared exponents, i.e., same as \texttt{FP32} and \texttt{bfloat16}, 
making the dynamic range of \texttt{FB12} wide enough to train DNNs.
As emphasized by the prior work, the precision and/or dynamic range during the weight gradient computation is extremely important
for the reliable DNN training~\cite{mixed_precision, flexpoint, hybrid-bfp, hfp8}.
Thus, we may elevate the bit precision to \texttt{FB16} during the weight update when training with \texttt{FB12}.
All computations use 4-bit except when computing the weight gradients (marked as \texttt{FB12+WG16}).
As shown in Fig.~\ref{fig:accuracy_imprv}, the test accuracy on CIFAR datasets mostly matches with the \texttt{FP32} baseline by using \texttt{FB12+WG16} in FlexBlock.
For DenseNet-121 on CIFAR-100, the accuracy with \texttt{FB12+WG16} is 2.31\% short from the \texttt{FP32} baseline
(but, still 4.4\% better than the model trained with \texttt{FB12}).
%Thus, it is better to train DenseNet-121 with \texttt{FB16} (or \texttt{FB12+WG24}) to keep the accuracy at par.
As shown in Fig.~\ref{fig:accuracy_new}, training ResNet-18 on ImageNet fails when we use \texttt{FB12+WG16}.
By elevating the precision to \texttt{FB12+WG24}, we can achieve similar accuracy to the baseline.
This set of experiments shows that supporting multi-mode BFP arithmetic maximizes the efficiency of DNN training.
%\todo{Jaeha}{add texts on ImageNet and WMT14 training after all data points are ready.}

\begin{table*}[]
\centering
\caption{Architectural comparisons between TPU-like systolic array, BitFusion-like BFP accelerator, and FlexBlock Cores \\in terms of area, power consumption, and energy efficiency}
\label{tab:hw_comparison}
\scalebox{0.81}{
\begin{tabular}{|c|c|c|c|}
\hline
\textbf{Training Hardware}          & \textbf{TPU-like Systolic Array (SA)}    & \textbf{BitFusion-based BFP Accelerator (BF)}      & \textbf{FlexBlock Cores (Proposed; FB)}                \\ \hline\hline
\textbf{Technology}                        & 65nm              & 65nm     &65nm                               \\ \hline
\textbf{Supported Precision}                      & \texttt{bfloat16}              & \texttt{FB12}, \texttt{FB16}, \texttt{FB24}               & \texttt{FB12}, \texttt{FB16}, \texttt{FB24}                            \\ \hline
\textbf{Array Size}                      & 128$\times$128         & 64$\times$64 (for \texttt{FB24})         & 54$\times$64 (for \texttt{FB24})               \\ \hline
\textbf{\# of Multipliers}                         & \textbf{128$\times$128}             & 256$\times$256 (\texttt{FB12}) / \textbf{128$\times$128 (\texttt{FB16})} / 64$\times$64 (\texttt{FB24}) & 216$\times$256 (\texttt{FB12}) / \textbf{108$\times$128 (\texttt{FB16})} / 54$\times$64 (\texttt{FB24}) \\ \hline
\textbf{Area {[}mm\textsuperscript{2}{]}}  & 74.38       & 40.22                                      & 33.82                                     \\ \hline
\textbf{Power Consumption {[}W{]}}        & 9.84        & 17.83 (\texttt{FB12}) / 17.13 (\texttt{FB16}) / 15.96 (\texttt{FB24})                                       & 8.27 (\texttt{FB12}) / 7.80 (\texttt{FB16}) / 7.36 (\texttt{FB24}) \\ \hline
\textbf{Clock Frequency}                   & 333MHz            & 333MHz                                       & 333MHz                                       \\ \hline
\textbf{Throughput {[}TFLOPS{]}} &\textbf{1.77}             & 2.82 (\texttt{FB12}) / \textbf{1.77 (\texttt{FB16})} / 0.66 (\texttt{FB24})                                          & 8.78 (\texttt{FB12}) / \textbf{3.35 (\texttt{FB16})} / 0.95 (\texttt{FB24}) \\ \hline
\textbf{Efficiency {[}GFLOPS/W{]}}  & \textbf{179.6}            & 157.96 (\texttt{FB12}) / \textbf{103.16 (\texttt{FB16})} / 41.11 (\texttt{FB24})                                          & 1,061.7 (\texttt{FB12}) / \textbf{428.9 (\texttt{FB16})} / 128.7 (\texttt{FB24})                                          \\ \hline
%\textbf{GFLOPS/(W$\cdot$mm\textsuperscript{2})}  & \textbf{0.13}                              & 0.43 (\texttt{FB12}) / \textbf{0.13 (\texttt{FB16})} / 0.014 (\texttt{FB24})                                          & 5.19 (\texttt{FB12}) / \textbf{1.34 (\texttt{FB16})} / 0.14 (\texttt{FB24})                                          \\ \hline
\end{tabular}}\vspace{0.5mm}
% Training Deep Neural Networks with 8-bit Floating Number, NIPS 2018: Multiplication (FP8) and Accumulation (FP16). The systolic array based on that paper.
% Other architectures have FP32 accumulation
\end{table*}

%The dynamic range is the log base 10 of the ratio of the largest to smallest 
%representable positive numbers.

\subsection{Area and Energy Analysis}\label{sec:analysis}

\begin{figure}[t]
    \centering
    \includegraphics[scale=0.44]{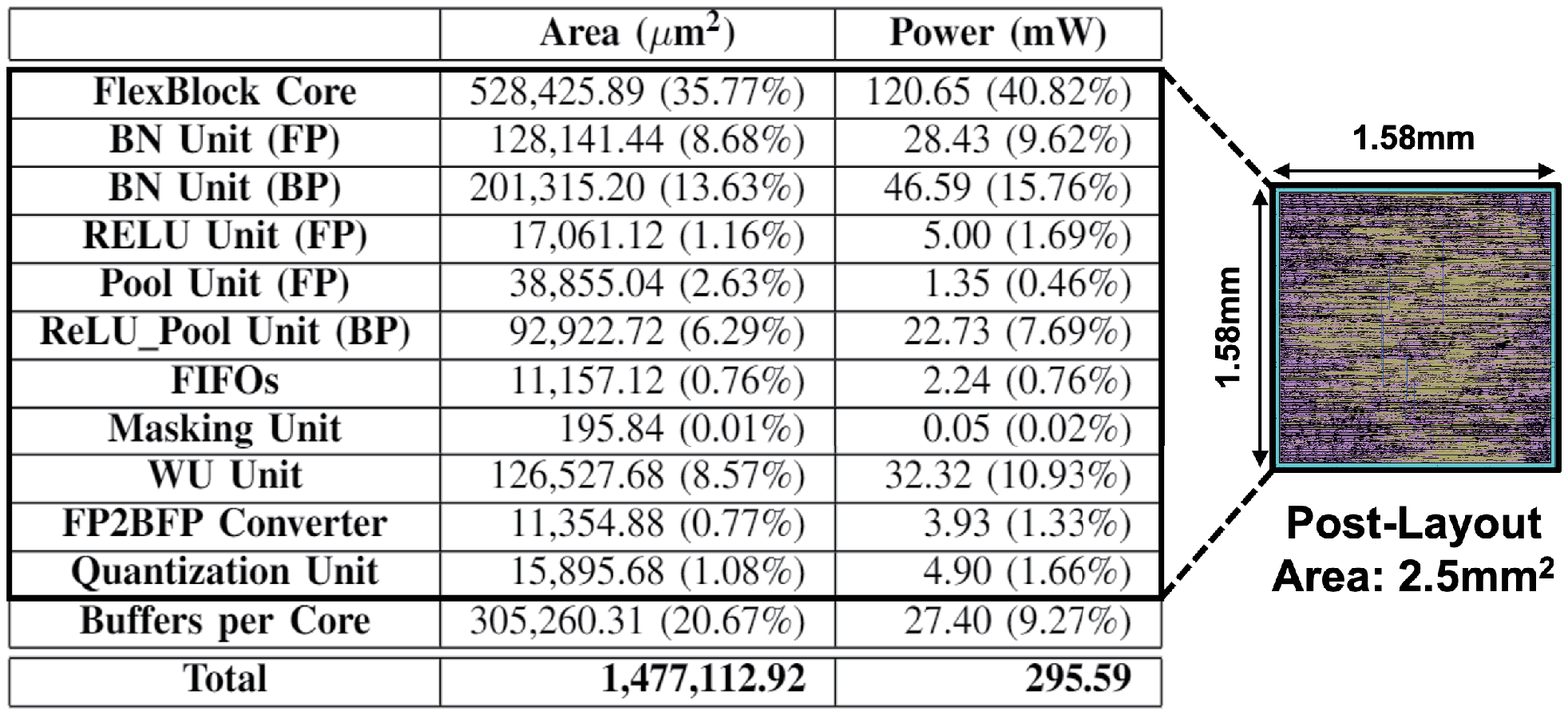}
    \caption{Area and power breakdowns of a FlexBlock core and hardware blocks for the BFP-based training.}
    \label{fig:area_power}\vspace{-1mm}
\end{figure}

To analyze the area and energy consumption,
we synthesized the RTL of a single FlexBlock core and all the required functional blocks
for the BFP-based training.
The reported area and power consumption are shown in Fig.~\ref{fig:area_power}.
The numbers for the FlexBlock core include the processing core, dual-path reduction units, and control blocks in Fig.~\ref{fig:overall_arch}.
About 36\% of area and 41\% of power consumption are used by the core.
%The BN units for FP and BP steps occupy 22\% of the total area 
%and consumes 24\% of the total power.
In total, 1.48mm\textsuperscript{2} of area ($\sim$2.81mm\textsuperscript{2} after PnR) and 295.59mW of power consumption are used by the single core.

\begin{figure*}[t]
    \centering
    \includegraphics[scale=0.84]{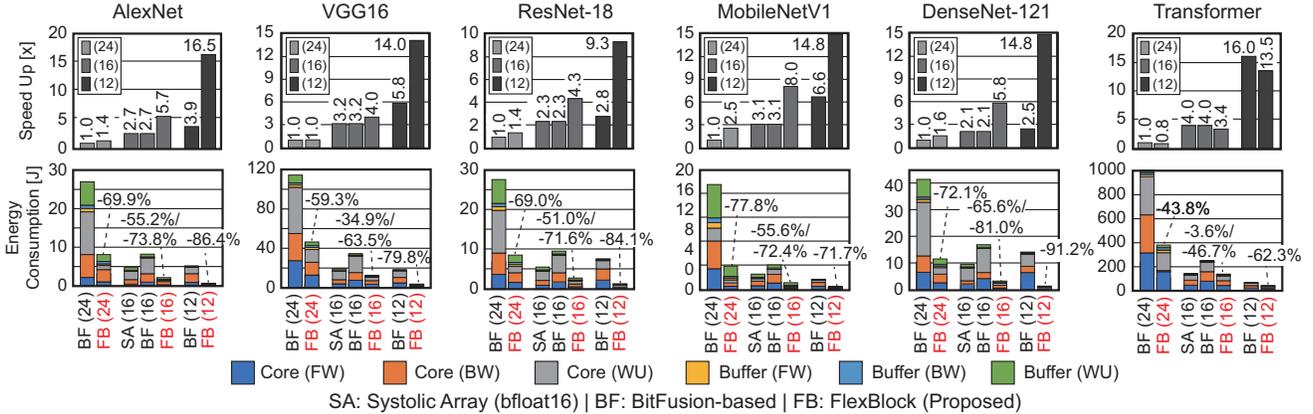}
    \caption{Comparisons of the performance and energy consumption between the systolic array (SA), the BitFusion-based BFP accelerator (BF), 
    and the proposed FlexBlock (FB in red).
    For the analysis, we evaluated five CNN benchmarks on ImageNet and Transformer-base model on WMT14. Here, \texttt{FB12} represents \texttt{FB12+WG24} format.} 
    \label{fig:energy_comp}
\end{figure*}

For more realistic analysis, we scaled up the FlexBlock accelerator with 64 cores,
which places 54$\times$64 full-precision (i.e., 16b$\times$16b) multipliers for the \texttt{FB24} mode.
RTLs of two baselines are designed and compared to FlexBlock in Table~\ref{tab:hw_comparison}: 
i) a systolic array using \texttt{bfloat16} (in short, SA) and ii) a BFP-based training accelerator 
using Bit Fusion architecture (in short, BF).
For BF, the array size is set to 64$\times$64 for the \texttt{FB24} mode.
The both FlexBlock and BF support multi-precision modes, e.g., \texttt{FB12}, \texttt{FB16} and \texttt{FB24}.
The array size of SA is set to 128$\times$128 to match the number of multipliers to the \texttt{FB16} mode
in BF or FlexBlock (i.e., 8-bit mantissas + 8-bit shared exponents; a BFP version of \texttt{bfloat16}).
All three training accelerators are running at 333MHz in 65nm CMOS technology.
The area of 64 FlexBlock cores (33.82mm\textsuperscript{2}) is 2.2$\times$ and 1.2$\times$ smaller than SA and BF, respectively.
The power consumption of 64 FlexBlock cores, i.e., 7.48mW on average, is 1.3$\times$ and 2.5$\times$ lower than
SA and BF, respectively.

To compare the throughput of three training accelerators,
RTL simulations were performed to extract the MAC utilization depending on the layer type, precision mode,
and tensor dimensions.
The extracted MAC utilization is being used in our cycle-approximate simulator 
to estimate the required clock cycles considering the memory access latency and the on-chip buffer size.
%For DRAM, we assume a dual-channel DDR4-2133 with a burst length of 8.
Instead of using small CIFAR datasets for CNN benchmarks,
we used ImageNet for all CNN models (with mini-batch size of 128) to compare three architectures 
in terms of the performance and energy consumption.
For Transformer model, we also used the mini-batch size of 128.
With the estimated clock cycles and the extracted power consumption of each accelerator, we report and compare the performance and energy consumption in Fig.~\ref{fig:energy_comp}.
%The energy consumption of tensor buffers include the access energy and the static leakage energy.
The training accelerators at an equivalent precision level are compared, 
e.g., SA with \texttt{bfloat16} is compared to BF and FlexBlock in \texttt{FB16}.
As a result, FlexBlock reduces the energy consumption (training time) by 65.3\%, 68.2\%, and 79.3\% (32.0\%, 47.5\%, and 68.4\%) on average compared to BF
at \texttt{FB24}, \texttt{FB16}, and \texttt{FB12+WG24}, respectively.
Compared to SA, FlexBlock reduces the energy consumption and training time by 44.3\% and 52.7\% on average.
When we train DNN models with \texttt{FB12+WG24} in FlexBlock, 
we can reduce the energy consumption and training time by 76.4\% and 81.0\% on average compared to SA.

\begin{table}[]
\centering
\caption{Comparisons of the performance and energy efficiency between GPU (NVIDIA RTX3090) and FlexBlock when training CNN benchmarks on ImageNet}
\label{tab:gpu_comp}
\scalebox{0.8}{
\begin{tabular}{|c|c|r|r|r|r|r|}
\hline
\multicolumn{2}{|c|}{\textbf{DNN Benchmark}}                                                                                           & \multicolumn{1}{c|}{\textbf{AlexNet}} & \multicolumn{1}{c|}{\textbf{VGG16}} & \multicolumn{1}{c|}{\textbf{ResNet}} & \multicolumn{1}{c|}{\textbf{MobileNet}} & \multicolumn{1}{c|}{\textbf{DenseNet}} \\ \hline\hline
\multirow{3}{*}{\textbf{\begin{tabular}[c]{@{}c@{}}GPU\\ (\texttt{FP16}\\ +\texttt{FP32})\end{tabular}}}        & \textbf{Runtime {[}ms{]}} & 46.0                                  & 296.4                             & 71.4                                 & 65.9                                    & 214.0                                  \\ \cline{2-7} 
                                                                                              & \textbf{Power {[}W{]}}    & 207.7                                 & 326.7                             & 321.4                                & 322.7                                   & 336.2                                  \\ \cline{2-7} 
                                                                                              & \textbf{GFLOPS/W}         & 41.1                                  & 61.0                              & 36.3                                 & 9.8                                     & 15.5                                   \\ \hline
\multirow{3}{*}{\textbf{\begin{tabular}[c]{@{}c@{}}FlexBlock\\ (\texttt{FB16})\end{tabular}}}          & \textbf{Runtime {[}ms{]}} & 178.8                                 & 1372.3                            & 243.6                                & 68.0                                    & 296.1                                  \\ \cline{2-7} 
                                                                                              & \textbf{Power {[}W{]}}    & 19.0                                  & 19.0                              & 19.0                                 & 19.0                                    & 19.0                                   \\ \cline{2-7} 
                                                                                              & \textbf{GFLOPS/W}         & 115.5                                 & 226.4                             & 179.9                                & 160.9                                   & 197.9                                  \\ \hline
\multirow{3}{*}{\textbf{\begin{tabular}[c]{@{}c@{}}FlexBlock\\  (\texttt{FB12}\\ +\texttt{WG24})\end{tabular}}} & \textbf{Runtime {[}ms{]}} & 61.8                                  & 391.4                             & 114.1                                & 36.7                                    & 116.9                                  \\ \cline{2-7} 
                                                                                              & \textbf{Power {[}W{]}}    & 19.5                                  & 19.5                              & 19.5                                 & 19.5                                    & 19.5                                   \\ \cline{2-7} 
                                                                                              & \textbf{GFLOPS/W}         & 325.9                                 & 774.4                             & 374.8                                & 290.8                                   & 489.0                                  \\ \hline
\end{tabular}}\vspace{0.5mm}
* All functional units listed in Fig.~\ref{fig:area_power} are included in the power report.\vspace{-1mm}
\end{table}

\subsection{Performance Comparison with GPU}

In this subsection, we compare the training speed and energy efficiency with a high-end GPU card, i.e., NVIDIA RTX3090.
When training in GPU, we utilized the mixed precision training (\texttt{FP16+FP32}) presented in~\cite{mixed_precision}.
The runtime for a single training iteration on 128 batches in GPU is measured by a built-in function in Python.
The power consumption of running each CNN benchmark is measured by \texttt{nvidia-smi}.
The reported numbers are summarized in Table~\ref{tab:gpu_comp}.
As one RTX3090 card has 384 Tensor Cores, it is equivalent to 20,992 \texttt{FP16} multipliers.
Thus, we compare the performance with FlexBlock using \texttt{FB16} and \texttt{FB12+WG24}.
The performance of FlexBlock with \texttt{FB16} is 2.9$\times$ lower than GPU.
However, this may come from the $\sim$1/4 of GPU clock frequency used by the current FlexBlock hardware.
The training speed with \texttt{FB12+WG24} on AlexNet, VGG16 and ResNet-18 is 1.4$\times$ slower than GPU.
However, training in FlexBlock is 1.8$\times$ faster on MobileNetV1 and DenseNet-121 than GPU.
Considering the 15.5$\times$ lower power consumption, FlexBlock in \texttt{FB12+WG24} achieves similar training speed
with much higher energy efficiency (18.4$\times$) compared to the recent GPU.

\subsection{Case Study: Dynamic Precision Control}\label{sec:dynamic}

\begin{figure}[t]
    \centering
    \includegraphics[scale=0.74]{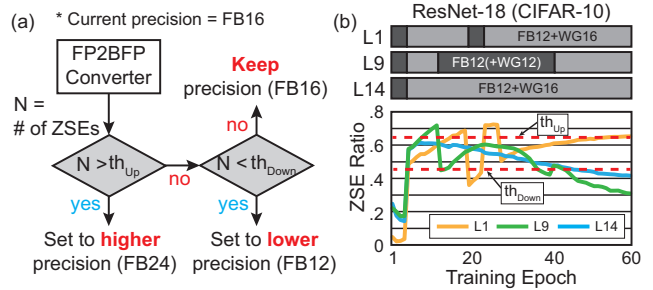}
    \caption{Dynamic precision control: (a) a hysteresis controller is used to determine the precision of a tensor for the next training epoch,
    (b) Layer-wise precision adaptation by checking the ZSE ratio (only three layers are shown for brevity).}
    \label{fig:dynamic}
\end{figure}

So far, we studied the benefit of statically assigning a different bit precision to each tensor for efficient DNN training.
However, it will be extremely useful if we can automatically tune the precision of each tensor at runtime while training a DNN model.
To accomplish this, we count the number of zero setting errors (ZSEs) due to the shift operations in the FP2BFP converter
explained in Fig.~\ref{fig:fp2bfp}.
We keep track of ZSEs of each tensor for the current epoch and determine the precision for the next training epoch by comparing the ratio of ZSEs to pre-defined thresholds.
We utilize a hysteresis controller to slowly change the bit precision (Fig.~\ref{fig:dynamic}(a)).
If the ratio of ZSEs is too large, it means that the current mantissa bit is not sufficient to train the model.
To demonstrate the feasibility of this approach, we fixed activation and weight precisions to \texttt{FB12}
and dynamically adjusted the precision of weight gradients between \texttt{FB12} and \texttt{FB16} at runtime.
We tested this approach on ResNet-18 with CIFAR-10 dataset.
Fig.~\ref{fig:fp2bfp}(b) shows how layer-wise precision adaptation is done by the proposed control mechanism.
Thanks to this dynamic precision control, we observed 16\% speed-up compared to the static \texttt{FB12+WG16} case with no accuracy degradation ($\sim$45\% of weight gradients, \texttt{WG}, were set to \texttt{FB12} instead of \texttt{FB16}).

\section{Related Work}\label{sec:related}

\subsection{Accelerators for Training Deep Neural Networks}

As training DNNs requires higher memory bandwidth and more computational resources
than the inference, many prior work proposed accelerators~\cite{deeptrain,scaledeep,gist}
or systems~\cite{dist_learning,fpdeep,gpipe,dpt} optimized for the training.
In ScaleDeep~\cite{scaledeep}, heterogeneous processing
tiles are utilized to map different types of DNN layers
for the efficient training.
In DeepTrain~\cite{deeptrain}, authors present temporally
heterogeneous tensor mapping with near-memory computing 
using a 3D-stacked memory.
Gist~\cite{gist} encodes the feature maps computed during the forward pass
to efficiently store them for later use in the backward pass.
%Feature maps are characterized by analyzing a pair of layers,
%i.e., ReLU-Pool or ReLU-Conv, and encoded differently
%Depending on the pair type, feature maps are binarized or encoded by a sparse data format
%to minimize the energy consumption in accessing DRAM.
In addition, many research focus on the distributed (or pipelined) training of DNN models
~\cite{dist_learning,fpdeep,gpipe,dpt} to achieve the best training performance.

Recently, sparse DNN accelerators are proposed
to increase the throughput of training DNNs by exploiting the possible sparsity
at each tensor~\cite{procrustes, sigma}.
SIGMA~\cite{sigma} proposes a training accelerator
that handles both sparsity and irregular structure in GEMM operations
by using a Benes network for efficient workload distribution.
Authors in~\cite{procrustes} present a sparse DNN training accelerator,
named Procrustes, that exploits one source of sparsity 
(either activations or weights)
during the forward pass, backward pass, or weight update.
Procrustes leverages the mini-batch dimension,
i.e., a dense tensor dimension, for the balanced workload distribution
when performing arithmetic operations involving sparse tensors.

\subsection{Reduced Precision During DNN Training}

To maximize the arithmetic density of training accelerators,
%To further reduce the complexity of floating-point arithmetic,
fixed-point logic can be used during the DNN training~\cite{gupta15, hybrid-bfp, flexpoint}.
In~\cite{gupta15}, stochastic rounding is used in training DNNs
with \texttt{INT16} to minimize the expected numerical error.
This work, however, evaluated the proposed method on relatively simple tasks,
i.e., classifying 10 different image classes using MNIST and CIFAR-10.
Other previous work aggressively reduce the precision during the training at the cost of noticeable accuracy degradation
~\cite{lin16,naveen17,hubara18,4bit_training,bnn,xnor-net,twn,alemdar16}.
To overcome the limited range of the fixed-point representation,
Flexpoint~\cite{flexpoint} extracts a 5-bit shared exponent for each tensor 
(coarse-grained) and adjusts the exponent twice per mini-batch
to prevent the overflows.
%to maximize the available dynamic range.
%To reduce the chance of numerical overflow,
To perform in-place exponent extraction, rather than 
periodically checking the overflow,
an accelerator with hybrid block floating point~\cite{hybrid-bfp} 
is proposed that performs multiply-and-accumulate operations on the
fixed-point logic while other remaining operations
are done in \texttt{FP32}.
%to maintain the model accuracy.
%Opposed to~\cite{flexpoint}, authors in~\cite{hybrid-bfp} tile
%the tensor with the size of 24$\times$24 at which each tile shares the exponent 
%providing better accuracy with lower bit-width.
Compared to~\cite{flexpoint, hybrid-bfp}, FlexBlock allows more fine-grained blocking
of sub-tensors to support variable precisions for accelerating the training process 
as discussed in Section~\ref{sec:analysis}.
%Other previous work aggressively reduce the precision
%of a tensor involved in the DNN training, 
%e.g., 8-bit~\cite{lin16,naveen17,rangeBN}, 4-bit~\cite{hubara18,4bit_training} or 
%even binary/ternary~\cite{bnn,xnor-net,twn,alemdar16}.
%However, noticeable accuracy degradation is inevitable with the aggressive precision reduction.
In the very recent work on low-precision training~\cite{fp8, seb_isscc, hfp8, hfp8_isscc, rapid}, 8-bit floating point (\texttt{FP8} or \texttt{HFP8}) 
has been used to train DNNs with a little accuracy loss on a wide spectrum of benchmarks.
However, the hardware associated with FP8 training uses specific mantissa and exponent bits 
for its maximum energy efficiency, which lacks flexibility.
%Moreover, some layers still require \texttt{FP16} computations such as depthwise convolution~\cite{hfp8}.

\begin{comment}
A trained model at a low precision, however, is only suited
for a DNN accelerator designed for that specific precision,
e.g., binary CNN with in-memory processing~\cite{xnor-pop} 
or 8-bit fixed-point on Google TPUv1~\cite{tpu}.
Instead, FlexBlock can train the model at various precision levels,
for both fixed-point (\texttt{INT4}, \texttt{INT8}, \texttt{INT16}) and 
floating point (\texttt{FB12}, \texttt{FB16}, \texttt{FB24}) representations, providing higher generality.
For instance, a DNN model trained in \texttt{FB12} or \texttt{FB16}
can be executed on an accelerator supporting
\texttt{bfloat16}~\cite{cloud-tpu,nnp-t} or CPEs/GPEs with single-precision (\texttt{FP32}; having the same exponent width).
A DNN model trained in \texttt{FB24} can be executed on CPEs/GPEs
with \texttt{FP32}.
\end{comment}

\section{Conclusion}\label{sec:conclusion}

In this work, we proposed a DNN training accelerator, i.e., FlexBlock, designed to support
multi-precision block floating point arithmetics.
This multi-mode BFP support has two main advantages: i) enabling users to train DNNs at desired precision levels,
and ii) reducing the training time for faster DNN exploration.
We identified the inherent limitation of the prior precision-scalable MAC arrays
and hierarchically allocated the tensor dimensions to compute units in FlexBlock for better performance.
%The tensor dimensions in deep learning models are delicately mapped to compute units within FlexBlock.
As the computations involved in the DNN training are rapidly increasing,
%and the wider deployment of DNNs in many applications,
this work will encourage developing training hardware with better flexibility
and higher energy efficiency using various BFP formats.

%\section*{Acknowledgements}
%This document is derived from previous conferences, in particular ISCA 2019,
%MICRO 2019, ISCA 2020, MICRO 2020, ISCA 2021, and HPCA 2022. We thank the
%organizers of HPCA 2022 for the idea to move the paper number and
%confidentiality banner to the \texttt{author} field to adhere to the HotCRP
%format checker.

%%%%%%% -- PAPER CONTENT ENDS -- %%%%%%%%

%%%%%%%%% -- BIB STYLE AND FILE -- %%%%%%%%
\bibliographystyle{IEEEtranS}
\bibliography{refs}

% Generated by IEEEtranS.bst, version: 1.13 (2008/09/30)
\begin{thebibliography}{10}
\providecommand{\url}[1]{#1}
\csname url@samestyle\endcsname
\providecommand{\newblock}{\relax}
\providecommand{\bibinfo}[2]{#2}
\providecommand{\BIBentrySTDinterwordspacing}{\spaceskip=0pt\relax}
\providecommand{\BIBentryALTinterwordstretchfactor}{4}
\providecommand{\BIBentryALTinterwordspacing}{\spaceskip=\fontdimen2\font plus
\BIBentryALTinterwordstretchfactor\fontdimen3\font minus
  \fontdimen4\font\relax}
\providecommand{\BIBforeignlanguage}[2]{{%
\expandafter\ifx\csname l@#1\endcsname\relax
\typeout{** WARNING: IEEEtranS.bst: No hyphenation pattern has been}%
\typeout{** loaded for the language `#1'. Using the pattern for}%
\typeout{** the default language instead.}%
\else
\language=\csname l@#1\endcsname
\fi
#2}}
\providecommand{\BIBdecl}{\relax}
\BIBdecl

\bibitem{hfp8_isscc}
A.~Agrawal, S.~K. Lee, J.~Silberman, M.~Ziegler, M.~Kang, S.~Venkataramani,
  N.~Cao, B.~Fleischer, M.~Guillorn, M.~Cohen, S.~Mueller, J.~Oh, M.~Lutz,
  J.~Jung, S.~Koswatta, C.~Zhou, V.~Zalani, J.~Bonanno, R.~Casatuta, C.-Y.
  Chen, J.~Choi, H.~Haynie, A.~Herbert, R.~Jain, M.~Kar, K.-H. Kim, Y.~Li,
  Z.~Ren, S.~Rider, M.~Schaal, K.~Schelm, M.~Scheuermann, X.~Sun, H.~Tran,
  N.~Wang, W.~Wang, X.~Zhang, V.~Shah, B.~Curran, V.~Srinivasan, P.-F. Lu,
  S.~Shukla, L.~Chang, and K.~Gopalakrishnan, ``9.1 a 7nm 4-core {AI} chip with
  25.6{TFLOPS} hybrid {FP8} training, {102.4TOPS INT4} inference and
  workload-aware throttling,'' in \emph{IEEE International Solid- State
  Circuits Conference (ISSCC)}, vol.~64, 2021, pp. 144--146.

\bibitem{alemdar16}
\BIBentryALTinterwordspacing
H.~Alemdar, N.~Caldwell, V.~Leroy, A.~Prost{-}Boucle, and F.~P{\'{e}}trot,
  ``Ternary neural networks for resource-efficient {AI} applications,''
  \emph{arXiv:1609.00222}, 2016. [Online]. Available:
  \url{http://arxiv.org/abs/1609.00222}
\BIBentrySTDinterwordspacing

\bibitem{robotics}
M.~Andrychowicz, B.~Baker, M.~Chociej, R.~Józefowicz, B.~McGrew, J.~Pachocki,
  A.~Petron, M.~Plappert, G.~Powell, A.~Ray, J.~Schneider, S.~Sidor, J.~Tobin,
  P.~Welinder, L.~Weng, and W.~Zaremba, ``Learning dexterous in-hand
  manipulation,'' \emph{The International Journal of Robotics Research (IJRR)},
  vol.~39, no.~1, pp. 3--20, 2020.

\bibitem{rangeBN}
R.~Banner, I.~Hubara, E.~Hoffer, and D.~Soudry, ``Scalable methods for 8-bit
  training of neural networks,'' in \emph{Proceedings of International
  Conference on Neural Information Processing Systems (NeurIPS)}, 2018, pp.
  5151--5159.

\bibitem{wmt14}
O.~Bojar, C.~Buck, C.~Federmann, B.~Haddow, P.~Koehn, J.~Leveling, C.~Monz,
  P.~Pecina, M.~Post, H.~Saint-Amand, R.~Soricut, L.~Specia, and A.~Tamchyna,
  ``Findings of the 2014 workshop on statistical machine translation,'' in
  \emph{Proceedings of the Ninth Workshop on Statistical Machine Translation
  (WMT)}, June 2014, pp. 12--58.

\bibitem{gpt3}
T.~B. Brown, B.~Mann, N.~Ryder, M.~Subbiah, J.~Kaplan, P.~Dhariwal,
  A.~Neelakantan, P.~Shyam, G.~Sastry, A.~Askell, S.~Agarwal, A.~Herbert-Voss,
  G.~Krueger, T.~Henighan, R.~Child, A.~Ramesh, D.~M. Ziegler, J.~Wu,
  C.~Winter, C.~Hesse, M.~Chen, E.~Sigler, M.~Litwin, S.~Gray, B.~Chess,
  J.~Clark, C.~Berner, S.~McCandlish, A.~Radford, I.~Sutskever, and D.~Amodei,
  ``Language models are few-shot learners,'' in \emph{Advances in Neural
  Information Processing Systems (NeurIPS)}, 2020.

\bibitem{survey19}
V.~Camus, L.~Mei, C.~Enz, and M.~Verhelst, ``Review and benchmarking of
  precision-scalable multiply-accumulate unit architectures for embedded
  neural-network processing,'' \emph{IEEE Journal on Emerging and Selected
  Topics in Circuits and Systems (JETCAS)}, vol.~9, no.~4, pp. 697--711, 2019.

\bibitem{pact}
\BIBentryALTinterwordspacing
J.~Choi, Z.~Wang, S.~Venkataramani, P.~I. Chuang, V.~Srinivasan, and
  K.~Gopalakrishnan, ``{PACT:} parameterized clipping activation for quantized
  neural networks,'' \emph{arXiv:1805.06085}, 2018. [Online]. Available:
  \url{http://arxiv.org/abs/1805.06085}
\BIBentrySTDinterwordspacing

\bibitem{imagenet}
J.~Deng, W.~Dong, R.~Socher, L.-J. Li, K.~Li, and L.~Fei-Fei, ``{ImageNet}: A
  large-scale hierarchical image database,'' in \emph{IEEE Conference on
  Computer Vision and Pattern Recognition (CVPR)}, 2009, pp. 248--255.

\bibitem{hybrid-bfp}
M.~Drumond, T.~Lin, M.~Jaggi, and B.~Falsafi, ``Training {DNNs} with hybrid
  block floating point,'' in \emph{Advances in Neural Information Processing
  Systems (NeurIPS)}, 2018.

\bibitem{fpdeep}
T.~Geng, T.~Wang, A.~Sanaullah, C.~Yang, R.~Xu, R.~Patel, and M.~Herbordt,
  ``{FPDeep}: Acceleration and load balancing of {CNN} training on {FPGA}
  clusters,'' in \emph{IEEE International Symposium on Field-Programmable
  Custom Computing Machines (FCCM)}, 2018, pp. 81--84.

\bibitem{cloud-tpu}
Google, ``Cloud {TPU},'' \url{https://cloud.google.com/tpu}, 2017, [Online;
  accessed 07-June-2021].

\bibitem{bfloat16}
{Google Cloud}, ``{BFloat16}: The secret to high performance on cloud {TPUs},''
  \url{https://cloud.google.com/blog/products/ai-machine-learning/bfloat16-the-secret-to-high-performance-on-cloud-tpus},
  2019, [Online; accessed 07-June-2021].

\bibitem{auto_driving}
\BIBentryALTinterwordspacing
S.~M. Grigorescu, B.~Trasnea, T.~T. Cocias, and G.~Macesanu, ``A survey of deep
  learning techniques for autonomous driving,'' \emph{arXiv:1910.07738}, 2019.
  [Online]. Available: \url{http://arxiv.org/abs/1910.07738}
\BIBentrySTDinterwordspacing

\bibitem{gupta15}
S.~Gupta, A.~Agrawal, K.~Gopalakrishnan, and P.~Narayanan, ``Deep learning with
  limited numerical precision,'' in \emph{Proceedings of International
  Conference on Machine Learning (ICML)}, vol.~37, July 2015, pp. 1737--1746.

\bibitem{han16}
\BIBentryALTinterwordspacing
S.~Han, H.~Mao, and W.~J. Dally, ``Deep compression: Compressing deep neural
  network with pruning, trained quantization and huffman coding,'' in
  \emph{International Conference on Learning Representations (ICLR)}, 2016.
  [Online]. Available: \url{http://arxiv.org/abs/1510.00149}
\BIBentrySTDinterwordspacing

\bibitem{resnet}
K.~He, X.~Zhang, S.~Ren, and J.~Sun, ``Deep residual learning for image
  recognition,'' in \emph{Proceedings of the IEEE Conference on Computer Vision
  and Pattern Recognition (CVPR)}, 2016, pp. 770--778.

\bibitem{nnp-t}
B.~Hickmann, J.~Chen, M.~Rotzin, A.~Yang, M.~Urbanski, and S.~Avancha, ``Intel
  {Nervana} neural network processor-t {(NNP-T)} fused floating point many-term
  dot product,'' in \emph{IEEE Symposium on Computer Arithmetic (ARITH)}, 2020,
  pp. 133--136.

\bibitem{mobilenet}
A.~G. Howard, M.~Zhu, B.~Chen, D.~Kalenichenko, W.~Wang, T.~Weyand,
  M.~Andreetto, and H.~Adam, ``Mobilenets: Efficient convolutional neural
  networks for mobile vision applications,'' \emph{arXiv preprint
  arXiv:1704.04861}, 2017.

\bibitem{densenet}
G.~Huang, Z.~Liu, G.~Pleiss, L.~Van Der~Maaten, and K.~Weinberger,
  ``Convolutional networks with dense connectivity,'' \emph{IEEE Transactions
  on Pattern Analysis and Machine Intelligence (TPAMI)}, pp. 1--1, 2019.

\bibitem{gpipe}
\BIBentryALTinterwordspacing
Y.~Huang, Y.~Cheng, D.~Chen, H.~Lee, J.~Ngiam, Q.~V. Le, and Z.~Chen,
  ``{GPipe}: Efficient training of giant neural networks using pipeline
  parallelism,'' \emph{arXiv:1811.06965}, 2018. [Online]. Available:
  \url{http://arxiv.org/abs/1811.06965}
\BIBentrySTDinterwordspacing

\bibitem{bnn}
I.~Hubara, M.~Courbariaux, D.~Soudry, R.~El-Yaniv, and Y.~Bengio, ``Binarized
  neural networks,'' in \emph{Proceedings of International Conference on Neural
  Information Processing Systems (NIPS)}, 2016, pp. 4114--4122.

\bibitem{hubara18}
I.~Hubara, M.~Courbariaux, D.~Soudry, R.~El-Yaniv, and Y.~Bengio, ``Quantized
  neural networks: Training neural networks with low precision weights and
  activations,'' \emph{Journal of Machine Learning Research (JMLR)}, vol.~18,
  no. 187, pp. 1--30, 2018.

\bibitem{tradBN}
S.~Ioffe and C.~Szegedy, ``Batch normalization: Accelerating deep network
  training by reducing internal covariate shift,'' in \emph{Proceedings of
  International Conference on International Conference on Machine Learning
  (ICML)}, 2015, pp. 448--456.

\bibitem{gist}
A.~Jain, A.~Phanishayee, J.~Mars, L.~Tang, and G.~Pekhimenko, ``Gist: Efficient
  data encoding for deep neural network training,'' in \emph{Proceedings of
  International Symposium on Computer Architecture (ISCA)}, 2018, pp. 776--789.

\bibitem{dpt}
Y.~Jang, S.~Kim, D.~Kim, S.~Lee, and J.~Kung, ``Deep partitioned training from
  near-storage computing to {DNN} accelerators,'' \emph{IEEE Computer
  Architecture Letters (CAL)}, pp. 1--4, 2021.

\bibitem{tpu}
N.~P. Jouppi, C.~Young, N.~Patil, D.~Patterson, G.~Agrawal, R.~Bajwa, S.~Bates,
  S.~Bhatia, N.~Boden, A.~Borchers, R.~Boyle, P.-l. Cantin, C.~Chao, C.~Clark,
  J.~Coriell, M.~Daley, M.~Dau, J.~Dean, B.~Gelb, T.~V. Ghaemmaghami,
  R.~Gottipati, W.~Gulland, R.~Hagmann, C.~R. Ho, D.~Hogberg, J.~Hu, R.~Hundt,
  D.~Hurt, J.~Ibarz, A.~Jaffey, A.~Jaworski, A.~Kaplan, H.~Khaitan,
  D.~Killebrew, A.~Koch, N.~Kumar, S.~Lacy, J.~Laudon, J.~Law, D.~Le, C.~Leary,
  Z.~Liu, K.~Lucke, A.~Lundin, G.~MacKean, A.~Maggiore, M.~Mahony, K.~Miller,
  R.~Nagarajan, R.~Narayanaswami, R.~Ni, K.~Nix, T.~Norrie, M.~Omernick,
  N.~Penukonda, A.~Phelps, J.~Ross, M.~Ross, A.~Salek, E.~Samadiani, C.~Severn,
  G.~Sizikov, M.~Snelham, J.~Souter, D.~Steinberg, A.~Swing, M.~Tan,
  G.~Thorson, B.~Tian, H.~Toma, E.~Tuttle, V.~Vasudevan, R.~Walter, W.~Wang,
  E.~Wilcox, and D.~H. Yoon, ``In-datacenter performance analysis of a tensor
  processing unit,'' \emph{SIGARCH Comput. Archit. News}, vol.~45, no.~2, pp.
  1--12, 2017.

\bibitem{bn2019}
W.~Jung, D.~Jung, B.~Kim, S.~Lee, W.~Rhee, and J.~H. Ahn, ``Restructuring batch
  normalization to accelerate {CNN} training,'' in \emph{Proceedings of the
  Conference on Systems and Machine Learning (SysML)}, 2019, pp. 1--13.

\bibitem{bfp_system}
K.~Kalliojarvi and J.~Astola, ``Roundoff errors in block-floating-point
  systems,'' \emph{IEEE Transactions on Signal Processing (TSP)}, vol.~44,
  no.~4, pp. 783--790, 1996.

\bibitem{deeptrain}
D.~Kim, T.~Na, S.~Yalamanchili, and S.~Mukhopadhyay, ``{DeepTrain}: A
  programmable embedded platform for training deep neural networks,''
  \emph{IEEE Transactions on Computer-Aided Design of Integrated Circuits and
  Systems (TCAD)}, vol.~37, no.~11, pp. 2360--2370, 2018.

\bibitem{flexpoint}
U.~K\"{o}ster, T.~J. Webb, X.~Wang, M.~Nassar, A.~K. Bansal, W.~H. Constable,
  O.~H. Elibol, S.~Gray, S.~Hall, L.~Hornof, A.~Khosrowshahi, C.~Kloss, R.~J.
  Pai, and N.~Rao, ``Flexpoint: An adaptive numerical format for efficient
  training of deep neural networks,'' in \emph{Proceedings of International
  Conference on Neural Information Processing Systems (NIPS)}, 2017, pp.
  1740--1750.

\bibitem{cifar-10}
A.~Krizhevsky, V.~Nair, and G.~Hinton, ``{CIFAR}-10 and {CIFAR}-100 dataset,''
  \url{https://www.cs.toronto.edu/~kriz/cifar.html}, 2010, [Online; accessed
  20-July-2021].

\bibitem{alexnet}
A.~Krizhevsky, I.~Sutskever, and G.~E. Hinton, ``Imagenet classification with
  deep convolutional neural networks,'' \emph{Advances in Neural Information
  Processing Systems (NIPS)}, vol.~25, pp. 1097--1105, 2012.

\bibitem{kung16}
J.~Kung, D.~Kim, and S.~Mukhopadhyay, ``Dynamic approximation with feedback
  control for energy-efficient recurrent neural network hardware,'' in
  \emph{Proceedings of the International Symposium on Low Power Electronics and
  Design (ISLPED)}, 2016, pp. 168--173.

\bibitem{unpu}
J.~Lee, C.~Kim, S.~Kang, D.~Shin, S.~Kim, and H.-J. Yoo, ``{UNPU}: A
  50.6{TOPS/W} unified deep neural network accelerator with 1b-to-16b
  fully-variable weight bit-precision,'' in \emph{IEEE International
  Solid-State Circuits Conference (ISSCC)}, 2018, pp. 218--220.

\bibitem{lnpu}
J.~Lee, J.~Lee, D.~Han, J.~Lee, G.~Park, and H.-J. Yoo, ``{LNPU}: A
  25.3{TFLOPS/W} sparse deep-neural-network learning processor with
  fine-grained mixed precision of {FP8-FP16},'' in \emph{IEEE International
  Solid- State Circuits Conference (ISSCC)}, 2019, pp. 142--144.

\bibitem{twn}
\BIBentryALTinterwordspacing
F.~Li and B.~Liu, ``Ternary weight networks,'' \emph{arXiv:1605.04711}, 2016.
  [Online]. Available: \url{http://arxiv.org/abs/1605.04711}
\BIBentrySTDinterwordspacing

\bibitem{cactip}
S.~Li, K.~Chen, J.~H. Ahn, J.~B. Brockman, and N.~P. Jouppi, ``{CACTI-P}:
  Architecture-level modeling for {SRAM}-based structures with advanced leakage
  reduction techniques,'' in \emph{IEEE/ACM International Conference on
  Computer-Aided Design (ICCAD)}, 2011, pp. 694--701.

\bibitem{lin16}
D.~D. Lin, S.~S. Talathi, and V.~S. Annapureddy, ``Fixed point quantization of
  deep convolutional networks,'' in \emph{Proceedings of International
  Conference on Machine Learning (ICML)}, 2016, pp. 2849--2858.

\bibitem{mcunet}
J.~Lin, W.-M. Chen, Y.~Lin, J.~Cohn, C.~Gan, and S.~Han, ``{MCUNet}: Tiny deep
  learning on {IoT} devices,'' in \emph{Advances in Neural Information
  Processing Systems (NeurIPS)}, vol.~33.\hskip 1em plus 0.5em minus
  0.4em\relax Curran Associates, Inc., 2020, pp. 11\,711--11\,722.

\bibitem{lin18_autonomous}
S.-C. Lin, Y.~Zhang, C.-H. Hsu, M.~Skach, M.~E. Haque, L.~Tang, and J.~Mars,
  ``The architectural implications of autonomous driving: Constraints and
  acceleration,'' in \emph{Proceedings of the International Conference on
  Architectural Support for Programming Languages and Operating Systems
  (ASPLOS)}, 2018, pp. 751--766.

\bibitem{naveen17}
\BIBentryALTinterwordspacing
N.~Mellempudi, A.~Kundu, D.~Das, D.~Mudigere, and B.~Kaul, ``Mixed
  low-precision deep learning inference using dynamic fixed point,''
  \emph{arXiv:1701.08978}, 2017. [Online]. Available:
  \url{http://arxiv.org/abs/1701.08978}
\BIBentrySTDinterwordspacing

\bibitem{mixed_precision}
\BIBentryALTinterwordspacing
P.~Micikevicius, S.~Narang, J.~Alben, G.~F. Diamos, E.~Elsen, D.~Garc{\'{\i}}a,
  B.~Ginsburg, M.~Houston, O.~Kuchaiev, G.~Venkatesh, and H.~Wu, ``Mixed
  precision training,'' \emph{arXiv:1710.03740}, 2017. [Online]. Available:
  \url{http://arxiv.org/abs/1710.03740}
\BIBentrySTDinterwordspacing

\bibitem{chip_design}
\BIBentryALTinterwordspacing
A.~Mirhoseini, A.~Goldie, M.~Yazgan, J.~W. Jiang, E.~Songhori, S.~Wang, Y.-J.
  Lee, E.~Johnson, O.~Pathak, A.~Nazi, J.~Pak, A.~Tong, K.~Srinivasa, W.~Hang,
  E.~Tuncer, Q.~V. Le, J.~Laudon, R.~Ho, R.~Carpenter, and J.~Dean, ``A graph
  placement methodology for fast chip design,'' \emph{Nature}, vol. 594, pp.
  207--212, June 2021. [Online]. Available:
  \url{https://doi.org/10.1038/s41586-021-03544-w}
\BIBentrySTDinterwordspacing

\bibitem{wacv16}
B.~Moons, B.~D. Brabandere, L.~V. Gool, and M.~Verhelst, ``Energy-efficient
  convnets through approximate computing,'' in \emph{IEEE Winter Conference on
  Applications of Computer Vision (WACV)}, Mar. 2016.

\bibitem{envision}
B.~Moons, R.~Uytterhoeven, W.~Dehaene, and M.~Verhelst, ``Envision: A
  0.26-to-10tops/w subword-parallel dynamic-voltage-accuracy-frequency-scalable
  convolutional neural network processor in 28nm {FDSOI},'' in \emph{IEEE
  International Solid-State Circuits Conference (ISSCC)}, 2017, pp. 246--247.

\bibitem{seb_isscc}
J.~Park, S.~Lee, and D.~Jeon, ``A 40nm {4.81TFLOPS/W} 8b floating-point
  training processor for non-sparse neural networks using shared exponent bias
  and 24-way fused multiply-add tree,'' in \emph{IEEE International Solid-
  State Circuits Conference (ISSCC)}, vol.~64, 2021, pp. 1--3.

\bibitem{sigma}
E.~{Qin}, A.~{Samajdar}, H.~{Kwon}, V.~{Nadella}, S.~{Srinivasan}, D.~{Das},
  B.~{Kaul}, and T.~{Krishna}, ``{SIGMA}: A sparse and irregular {GEMM}
  accelerator with flexible interconnects for {DNN} training,'' in \emph{IEEE
  International Symposium on High Performance Computer Architecture (HPCA)},
  Feb. 2020, pp. 58--70.

\bibitem{xnor-net}
M.~Rastegari, V.~Ordonez, J.~Redmon, and A.~Farhadi, ``{XNOR-Net: ImageNet}
  classification using binary convolutional neural networks,'' in
  \emph{European Conference on Computer Vision (ECCV)}.\hskip 1em plus 0.5em
  minus 0.4em\relax Springer International Publishing, 2016, pp. 525--542.

\bibitem{amoebanet}
\BIBentryALTinterwordspacing
E.~Real, A.~Aggarwal, Y.~Huang, and Q.~V. Le, ``Regularized evolution for image
  classifier architecture search,'' \emph{arXiv:1802.01548}, 2018. [Online].
  Available: \url{http://arxiv.org/abs/1802.01548}
\BIBentrySTDinterwordspacing

\bibitem{yolo}
J.~Redmon, S.~Divvala, R.~Girshick, and A.~Farhadi, ``You only look once:
  Unified, real-time object detection,'' \emph{arXiv:1506.02640}, 2016.

\bibitem{bitblade}
S.~Ryu, H.~Kim, W.~Yi, and J.-J. Kim, ``{BitBlade}: Area and energy-efficient
  precision-scalable neural network accelerator with bitwise summation,'' in
  \emph{ACM/IEEE Design Automation Conference (DAC)}, 2019, pp. 1--6.

\bibitem{mobilenetv2}
M.~Sandler, A.~Howard, M.~Zhu, A.~Zhmoginov, and L.-C. Chen, ``{MobileNetV2}:
  Inverted residuals and linear bottlenecks,'' in \emph{IEEE/CVF Conference on
  Computer Vision and Pattern Recognition (CVPR)}, 2018, pp. 4510--4520.

\bibitem{loom}
S.~Sharify, A.~D. Lascorz, K.~Siu, P.~Judd, and A.~Moshovos, ``Loom: Exploiting
  weight and activation precisions to accelerate convolutional neural
  networks,'' in \emph{ACM/ESDA/IEEE Design Automation Conference (DAC)}, 2018,
  pp. 1--6.

\bibitem{bitfusion}
H.~Sharma, J.~Park, N.~Suda, L.~Lai, B.~Chau, V.~Chandra, and H.~Esmaeilzadeh,
  ``{Bit Fusion}: Bit-level dynamically composable architecture for
  accelerating deep neural network,'' in \emph{ACM/IEEE International Symposium
  on Computer Architecture (ISCA)}, 2018, pp. 764--775.

\bibitem{dnpu}
D.~Shin, J.~Lee, J.~Lee, and H.-J. Yoo, ``{DNPU}: An 8.1{TOPS/W} reconfigurable
  {CNN-RNN} processor for general-purpose deep neural networks,'' in \emph{IEEE
  International Solid-State Circuits Conference (ISSCC)}, 2017, pp. 240--241.

\bibitem{vggnet}
K.~Simonyan and A.~Zisserman, ``Very deep convolutional networks for
  large-scale image recognition,'' \emph{arXiv preprint arXiv:1409.1556}, 2014.

\bibitem{hfp8}
X.~Sun, J.~Choi, C.-Y. Chen, N.~Wang, S.~Venkataramani, V.~V. Srinivasan,
  X.~Cui, W.~Zhang, and K.~Gopalakrishnan, ``Hybrid 8-bit floating point (hfp8)
  training and inference for deep neural networks,'' in \emph{Advances in
  Neural Information Processing Systems (NeurIPS)}, H.~Wallach, H.~Larochelle,
  A.~Beygelzimer, F.~d\textquotesingle Alch\'{e}-Buc, E.~Fox, and R.~Garnett,
  Eds., vol.~32, 2019.

\bibitem{4bit_training}
X.~Sun, N.~Wang, C.-Y. Chen, J.~Ni, A.~Agrawal, X.~Cui, S.~Venkataramani,
  K.~El~Maghraoui, V.~V. Srinivasan, and K.~Gopalakrishnan, ``Ultra-low
  precision 4-bit training of deep neural networks,'' in \emph{Advances in
  Neural Information Processing Systems (NeurIPS)}, vol.~33, 2020, pp.
  1796--1807.

\bibitem{synopsys_icc}
Synopsys, ``{IC Compiler II Implementation User Guide: Version L-2016.03},''
  2016.

\bibitem{synopsys_dc}
Synopsys, ``{Design Compiler User Guide: Version N-2017.09},'' 2017.

\bibitem{effnet}
M.~Tan and Q.~Le, ``{E}fficient{N}et: Rethinking model scaling for
  convolutional neural networks,'' in \emph{Proceedings of the International
  Conference on Machine Learning (ICML)}, vol.~97, June 2019, pp. 6105--6114.

\bibitem{transformer}
A.~Vaswani, N.~Shazeer, N.~Parmar, J.~Uszkoreit, L.~Jones, A.~N. Gomez, L.~u.
  Kaiser, and I.~Polosukhin, ``Attention is all you need,'' in \emph{Advances
  in Neural Information Processing Systems (NeurIPS)}, vol.~30, 2017.

\bibitem{scaledeep}
S.~Venkataramani, A.~Ranjan, S.~Banerjee, D.~Das, S.~Avancha, A.~Jagannathan,
  A.~Durg, D.~Nagaraj, B.~Kaul, P.~Dubey, and A.~Raghunathan, ``Scaledeep: A
  scalable compute architecture for learning and evaluating deep networks,'' in
  \emph{ACM/IEEE International Symposium on Computer Architecture (ISCA)},
  2017, pp. 13--26.

\bibitem{rapid}
S.~Venkataramani, V.~Srinivasan, W.~Wang, S.~Sen, J.~Zhang, A.~Agrawal, M.~Kar,
  S.~Jain, A.~Mannari, H.~Tran, Y.~Li, E.~Ogawa, K.~Ishizaki, H.~Inoue,
  M.~Schaal, M.~Serrano, J.~Choi, X.~Sun, N.~Wang, C.-Y. Chen, A.~Allain,
  J.~Bonano, N.~Cao, R.~Casatuta, M.~Cohen, B.~Fleischer, M.~Guillorn,
  H.~Haynie, J.~Jung, M.~Kang, K.-h. Kim, S.~Koswatta, S.~Lee, M.~Lutz,
  S.~Mueller, J.~Oh, A.~Ranjan, Z.~Ren, S.~Rider, K.~Schelm, M.~Scheuermann,
  J.~Silberman, J.~Yang, V.~Zalani, X.~Zhang, C.~Zhou, M.~Ziegler, V.~Shah,
  M.~Ohara, P.-F. Lu, B.~Curran, S.~Shukla, L.~Chang, and K.~Gopalakrishnan,
  ``{RaPiD}: {AI} accelerator for ultra-low precision training and inference,''
  in \emph{ACM/IEEE International Symposium on Computer Architecture (ISCA)},
  2021, pp. 153--166.

\bibitem{dist_learning}
\BIBentryALTinterwordspacing
J.~Verbraeken, M.~Wolting, J.~Katzy, J.~Kloppenburg, T.~Verbelen, and J.~S.
  Rellermeyer, ``A survey on distributed machine learning,'' \emph{ACM Comput.
  Surv.}, vol.~53, no.~2, Mar. 2020. [Online]. Available:
  \url{https://doi.org/10.1145/3377454}
\BIBentrySTDinterwordspacing

\bibitem{fp8}
N.~Wang, J.~Choi, D.~Brand, C.-Y. Chen, and K.~Gopalakrishnan, ``Training deep
  neural networks with 8-bit floating point numbers,'' in \emph{Advances in
  Neural Information Processing Systems (NeurIPS)}, 2018, p. 7686–7695.

\bibitem{cacti}
S.~Wilton and N.~Jouppi, ``{CACTI}: an enhanced cache access and cycle time
  model,'' \emph{IEEE Journal of Solid-State Circuits (JSSC)}, vol.~31, no.~5,
  pp. 677--688, 1996.

\bibitem{procrustes}
D.~Yang, A.~Ghasemazar, X.~Ren, M.~Golub, G.~Lemieux, and M.~Lis, ``Procrustes:
  a dataflow and accelerator for sparse deep neural network training,'' in
  \emph{Proceedings of the IEEE/ACM International Symposium on
  Microarchitecture (MICRO)}, Oct. 2020.

\bibitem{yoon21}
J.-H. Yoon, M.~Chang, W.-S. Khwa, Y.-D. Chih, M.-F. Chang, and A.~Raychowdhury,
  ``A {40nm 64Kb 56.67TOPS/W} read-disturb-tolerant compute-in-memory/digital
  {RRAM} macro with active-feedback-based read and in-situ write
  verification,'' in \emph{IEEE International Solid- State Circuits Conference
  (ISSCC)}, vol.~64, 2021, pp. 404--406.

\end{thebibliography}
%%%%%%%%%%%%%%%%%%%%%%%%%%%%%%%%%%%%

\end{document}